%% file: VQD_EMNLP2024.tex
\pdfoutput=1

\documentclass[11pt]{article}

\usepackage[final]{acl}

\usepackage{times}
\usepackage{latexsym}
\usepackage{makecell}
\usepackage{multirow}
\usepackage{booktabs} 

\usepackage[T1]{fontenc}

\usepackage[utf8]{inputenc}

\usepackage{microtype}

\usepackage{inconsolata}

\usepackage{graphicx}

\usepackage{amssymb}
\usepackage{amsmath}
\usepackage{subcaption}
\usepackage{booktabs}
\usepackage{todonotes}
\usepackage{enumitem}
\usepackage{arydshln}
\usepackage{color}
\usepackage{multirow}
\usepackage{comment}
\usepackage{makecell}
\usepackage{tcolorbox}
\usepackage{fontawesome5}
\usepackage[hang]{footmisc}
\usepackage[ruled,noend,vlined]{algorithm2e}

\newcommand{\evalFrame}{SubQuestRater\xspace}
\newcommand{\down}[1]{\textcolor{red}{#1}}
\newcommand{\up}[1]{\textcolor{green!55!black}{#1}}

\title{Visual Question Decomposition on Multimodal Large Language Models}

\author{Haowei Zhang\thanks{Equal Contribution.}$^{~1}$ \quad Jianzhe Liu$\footnotemark[1]^{~1}$ \quad Zhen Han\thanks{Corresponding authors: \\Zhen Han <hanzhen02111@163.com>, this work does not relate to his position at AWS; Jindong Gu <jindong.gu@outlook.com>}$^{~2}$ \quad Shuo Chen$^{~3}$\\ \textbf{\quad Bailan He$^{~3}$\quad\quad Volker Tresp$^{~3,4}$ \quad Zhiqiang Xu$^{~5}$\quad Jindong Gu$\footnotemark[2]^{~6}$} \vspace{4pt}\\
$^{1}$Technical University of Munich, $^{2}$Amazon Web Services, \\$^{3}$LMU Munich, $^{4}$Munich Center for Machine Learning,\\ $^{5}$MBZUAI, $^{6}$University of Oxford\vspace{4pt}\\
\texttt{\{haowei.zhang, jianzhe.liu\}@tum.de}
}

\begin{document}

\maketitle

\begin{minipage}[h]{2\linewidth}
\vspace{-1cm}
  \centering
  \href{https://vqd-emnlp2024.github.io/}{{\faGithub{}}\xspace\texttt{https://vqd-emnlp2024.github.io/}} \\\vspace{2pt}
\vspace{0.5cm}
\end{minipage}

\input{chapters/0-abstract}

\input{chapters/1-introduction}
\input{chapters/2-related_work}
\input{chapters/3-how_well_MLLMs_decompose}

\input{chapters/4-enhancing_VQD_ability}
\input{chapters/5-experiments}
\input{chapters/6-conclusion}
\input{chapters/limitation}
\section*{Acknolwedgement}
The authors acknowledge support by the German Federal Ministry for Education and Research (BMBF), funding project Software Campus 2.0 / C-R-KG (FKZ 01IS17048).
\bibliography{VQD_EMNLP2024}

\input{chapters/appendix}

\end{document}

%% file: chapters/0-abstract.tex
\begin{abstract}
Question decomposition has emerged as an effective strategy for prompting Large Language Models (LLMs) to answer complex questions. However, while existing methods primarily focus on unimodal language models, the question decomposition capability of Multimodal Large Language Models (MLLMs) has yet to be explored. To this end, this paper explores visual question decomposition on MLLMs. Specifically, we introduce a systematic evaluation framework including a dataset and several evaluation criteria to assess the quality of the decomposed sub-questions, revealing that existing MLLMs struggle to produce high-quality sub-questions. To address this limitation, we propose a specific finetuning dataset, DecoVQA+, for enhancing the model's question decomposition capability. Aiming at enabling models to perform appropriate selective decomposition, we propose an efficient finetuning pipeline. The finetuning pipeline consists of our proposed dataset and a training objective for selective decomposition. Finetuned MLLMs demonstrate significant improvements in the quality of sub-questions and the policy of selective question decomposition. Additionally, the models also achieve higher accuracy with selective decomposition on VQA benchmark datasets.
\end{abstract}

%% file: chapters/1-introduction.tex
\section{Introduction}
\label{sec:intro}






Answering complex questions is a challenging task, especially when the questions require implicit multi-step reasoning to answer. Question Decomposition (QD) is an effective strategy to address this issue.
Most related work studies the efficacy of QD with unimodal textual large language models (LLMs) in enhancing complex textual question answering tasks~\citep{patel-etal-2022-question, dua-etal-2022-successive, zhou2023leasttomost, qi-etal-2023-art}. Although some recent works~\citep{you-etal-2023-idealgpt, qi-etal-2023-art} have explored question decomposition within the context of visual question answering (VQA) tasks, they follow the paradigm of performing unimodal QD based on the image caption. Typically, they conduct a two-step process: first, generating a caption for the image using a captioning model, and then performing question decomposition using an unimodal textual LLM based on the complex question and the generated image caption. Relying solely on the image caption instead of the image itself may lead to significant information loss. 

Recent advancements in Multimodal Large Language Models (MLLMs) have enabled MLLMs to directly perceive image information for answering questions. Yet, how to perform QD on complex visual questions using such MLLMs has been less explored. In the following, we refer to question decomposition using MLLMs on VQA as Visual Question Decomposition (VQD).
In this work, we primarily explore the following research questions:
\begin{itemize}
    \item How can we quantitatively assess the VQD ability of MLLMs? How proficient are existing MLLMs in VQD, or specifically, how is the quality of sub-questions generated by MLLMs?
    \item How can we enhance the VQD ability of MLLMs and enable the models to properly determine when to decompose and when not to, facing questions with varying difficulties?
\end{itemize}

\begin{figure}[!htb] 
\centering 
\includegraphics[width=\linewidth]{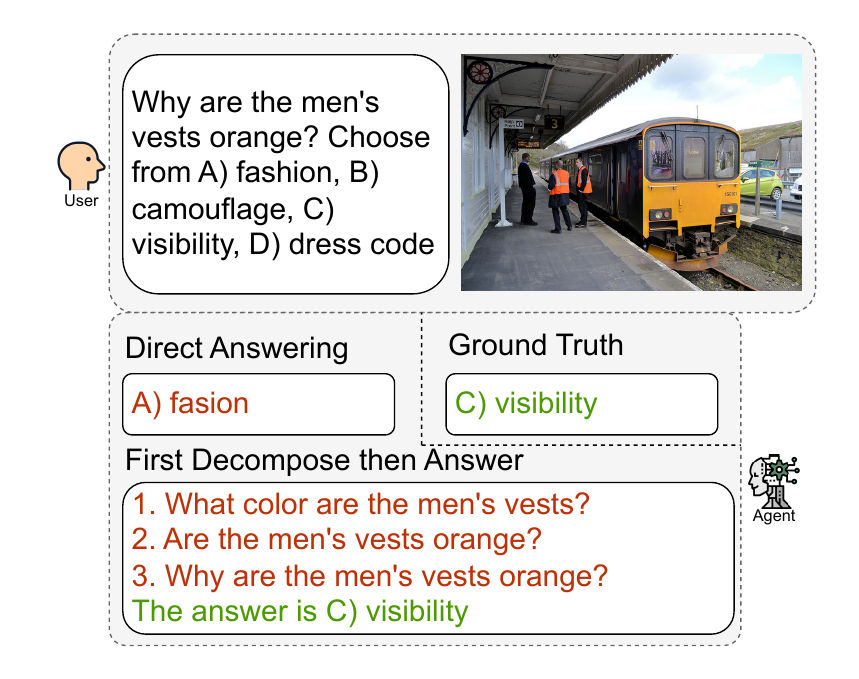}
\caption{Cases showing that even if the model correctly answers the original question, the generated sub-questions are of low quality: they are irrelevant or repeated from the original question.} 
\label{logo1} 
\end{figure}

  To assess the question decomposition capability of MLLMs, a significant obstacle is the absence of metrics for evaluating models' question decomposition abilities.
  Recent work~\citep{you-etal-2023-idealgpt,qi-etal-2023-art} evaluates the model's question decomposition ability by measuring the final answer's accuracy. However, relying solely on whether a model can correctly answer the original question is an implicit measure of its decomposition ability. 
  Especially, even if the final answer is correct, we have observed various issues in the decomposed sub-questions: for example, some MLLMs produce many repetitive sub-questions, or some sub-questions are entirely irrelevant to the original question, as shown in Figure~\ref{logo1}. 

  Ideally, the sub-questions should be highly relevant to the original question and not repetitive with the original question or other sub-questions. Besides, they should be relatively easy to be grounded, whose answer can be derived from images or pre-trained commonsense knowledge. Figure~\ref{fig2} shows a detailed comparison between sub-questions of high- and low-quality. To this end, we propose SubQuestRater, an evaluation framework for assessing MLLM's question decomposition ability. Specifically, considering the observed common deficiencies of existing MLLMs' question decomposition ability, we choose three critical criteria to assess the quality of question decomposition: 1) Non-Repetition, 2) Relevance, and 3) Groundedness.
 SubQuestRater quantifies the quality of each sub-question by assigning scores based on each criterion. 
 
 Besides, it is necessary to have an evaluation dataset containing complex questions requiring decomposition. However, current QA datasets, even specifically aiming at complex reasoning, such as A-OKVQA~\citep{schwenk2022aokvqa}, still contain a large number of simple questions that do not require decomposition to answer. Given the lack of publicly available benchmarks solely focusing on complex questions that require decomposition, we introduce a specific question decomposition evaluation dataset.
 With the help of our proposed evaluation criteria and benchmarks, we evaluate several MLLMs including MiniGPT-v2~\citep{chen2023minigptv2}, LLaVA-1.5~\citep{liu2024improved}, etc. The results show that, the decomposed sub-questions generated by these MLLMs cannot perform satisfactorily, demonstrating repetition, irrelevance, or ungroundedness in many cases.

  
  To enhance MLLMs' capability for VQD, we propose a new finetuning dataset tailored for question decomposition, DecoVQA. It is the first public dataset that consists of manually annotated sub-questions for complex questions. The provided sub-questions feature high quality in the view of non-repetition, relevance and groundedness. To prevent catastrophic forgetting, samples with simple questions in the form of direct answering are added to construct DecoVQA. We finetune MLLMs on this dataset with LoRA~\citep{hu2021lora}. 
  
  Furthermore, we find that existing MLLMs struggle to determine whether they need question decomposition to enhance their reasoning performance when facing problems of varying difficulty. To address this issue, we 
  propose a training pipeline with an upgraded version of DecoVQA, i.e. DecoVQA+, with an extra QA round asking models whether to decompose, and a novel objective function combining a next-token prediction loss (NTP loss) and a binary cross entropy loss (BCE loss) to fine-tune MLLMs. In addition to applying the conventional NTP loss for general reasoning, we design a BCE loss that aims to penalize the errors in deciding whether to decompose questions. Extensive experiments show that MLLMs after finetuning achieve a higher answer accuracy and learn to know when to decompose properly.

To summarize, the main contributions of our work are as follows:
\begin{itemize} [leftmargin=*]
\itemsep0em
\item[1.] We are the first to systematically investigate MLLMs' ability on visual question decomposition. We propose a comprehensive evaluation framework, SubQuestRater, which includes a benchmark dataset and novel evaluation metrics, to quantitatively evaluate the quality of generated sub-questions from diverse perspectives.

\item[2.] We find that existing MLLMs are insufficient to produce sub-questions with high quality. Efficient finetuning of MLLMs on our proposed dataset, DecoVQA, significantly improves their VQD ability. 

\item[3.] 
We propose a finetuning pipeline with an upgraded dataset, DecoVQA+, and a specific training objective for selective VQD, demonstrating improvements in the model's decision-making regarding decomposing questions or direct answering, as well as the accuracy of final answers.
\end{itemize}

%% file: chapters/2-related_work.tex
\section{Related Work}

\subsection{Question Decomposition}

Question decomposition has shown impressive capabilities in improving the reasoning performance of language models. Successive Prompting~\citep{dua-etal-2022-successive} and Least-to-Most Prompting~\citep{zhou2023leasttomost} are two representative works that break a complicated question into simpler ones iteratively. Decomposed Prompting~\citep{khot2023decomposed} introduces a modular setup of question decomposition, which makes it easy to optimize prompts, pretrained models and symbolic functions for different sub-tasks. Additionally, question decomposition is capable of increasing the reasoning faithfulness while achieving the accuracy improvement~\citep{radhakrishnan2023question}.


Recent studies have explored the potential of VQD. 
IdealGPT~\citep{you-etal-2023-idealgpt} leverages LLMs iteratively to raise sub-questions and determines the final reasoning answer. Socratic Questioning~\citep{qi-etal-2023-art} utilizes LLMs to generate sub-questions and answer them, stimulating robust recursive thinking. However, all of these studies rely on the reasoning ability of language models, overlooking the visual information that images can bring to question decomposition. The literature most relevant to our work is \cite{NEURIPS2023_b14cf0a0}, which explored prompting MLLMs to answer VQA questions with question decomposition in the zero-shot settings. However, it only applies VQD as a prompting technique and evaluates whether VQD could enhance the VQA accuracy. That work does not delve into the quality of generated sub-questions along the entire reasoning process, which does not explicitly analyze how well the questions are decomposed.

\subsection{Multimodal LLMs}
To address the modality gap, MiniGPT-v2~\citep{chen2023minigptv2} and LLaVA-1.5~\citep{liu2024improved} apply a linear connection layer to connect the frozen pre-trained vision module and the language model. Besides, they provide a task-oriented instruction training pipeline to decrease instructional ambiguity across various vision-language tasks. GPT-4 with Vision (GPT-4V)~\citep{openai2024gpt4} is a powerful MLLM based on instructing GPT-4 and has shown outstanding capability on diverse benchmarks. Besides instruction-following ability, Qwen-VL~\citep{bai2023qwenvl} series outperforms in a range of vision-language tasks, supporting multilingual conversations and dialogues involving multiple interleaved images. Furthermore, InternVL-1.5~\citep{chen2024far} introduces a strong vision encoder, dynamic high-resolution images, and a high-quality bilingual dataset to enhance the comprehensive capability of MLLMs further. While these models have demonstrated impressive performance across various benchmarks~\citep{bai2023qwenvl}, they continue to struggle with complex tasks that require advanced reasoning~\citep{NEURIPS2023_b14cf0a0}. Numerous techniques, such as parameter-efficient tuning methods like prompting~\citep{gu2023systematic}, in-context learning~\cite{alayrac2022flamingo}, and chain-of-thought reasoning~\citep{zhang2022automatic}, can be used to help models handle unseen or complex tasks, but each has its limitations. Parameter-efficient tuning is vulnerable to robustness issues when dealing with out-of-distribution inputs~\citep{chen2024benchmarking}, in-context learning often fails to fully utilize multimodal information, focusing predominantly on text~\citep{chen2023understanding}, and chain-of-thought reasoning is prone to adversarial attacks~\citep{wang2024stop}. In contrast, our approach employs question decomposition, breaking down complex queries into simpler, more manageable sub-questions, which enhances the models' reasoning capabilities.

%% file: chapters/3-how_well_MLLMs_decompose.tex
\section{How well can MLLMs decompose questions?}

Existing works commonly use the accuracy of the final answer to demonstrate a model's ability to decompose questions. However, this evaluation method is imprecise and implicit. As shown in Figure \ref{logo1}, a MLLM generates sub-questions with low quality, yet can still provide a correct answer. However, these ineffective sub-questions fail to provide the expected assistance in answering the original question and do not help the model's reasoning process. To address this, we differentiate the model's question decomposition skills from the accuracy of answering and propose SubQuestRater, an evaluation framework focusing explicitly on VQD ability. The framework consists of criteria that are specifically designed to assess the quality of sub-questions and an evaluation dataset.

To determine the criteria for the proposed framework, we have analyzed the generated sub-questions by existing MLLMs including MiniGPT-v2, LLaVA, etc., and we have observed the most common issues among them: Some sub-questions repeat the original question or other sub-questions (e.g., semantically equivalent), while others are not relevant to the original question. Besides, some sub-questions cannot be answered from images or commonsense knowledge. These issues largely influence the quality of sub-questions. After conducting the manual review and analysis of the sub-questions, we proposed three criteria to assess the quality of sub-questions, as follows:

\paragraph{Non-Repetition} This criterion ensures that sub-questions do not repeat. The definition of repetition here is the case where sub-questions discuss the same topic with the same or different phrasing. For example, in Figure~\ref{logo1}, the original question asks why the men's vests are orange, yet all the sub-questions repeatedly talk about orange vests, causing only redundancy.

\paragraph{Relevance} This criterion judges whether a sub-question truly contributes to answering the original question. For example, if the original question asks about the relationship between two people sitting at a table, but the sub-questions inquire about the colors of their clothes or the shapes of the table, these sub-questions are irrelevant. Such distractions can even mislead the model and reduce its performance.

\paragraph{Groundedness} This criterion evaluates whether a sub-question can be answered using information directly provided by the image or from commonsense knowledge. Given a relevant and not repeated sub-question, if it can't be grounded from image or commonsense knowledge, it would still be unhelpful.
For example, if the original question asks whether it is safe to cross the road now, and a sub-question inquires what is the time displayed on the traffic light, which helps answer the original question if it can be inferred from the image. However, the image only shows the yellow traffic light and there is no number on the light indicating the remaining time. Therefore, the sub-question is considered ungrounded.

\begin{figure}[h]
\centering
\includegraphics[width=\linewidth]{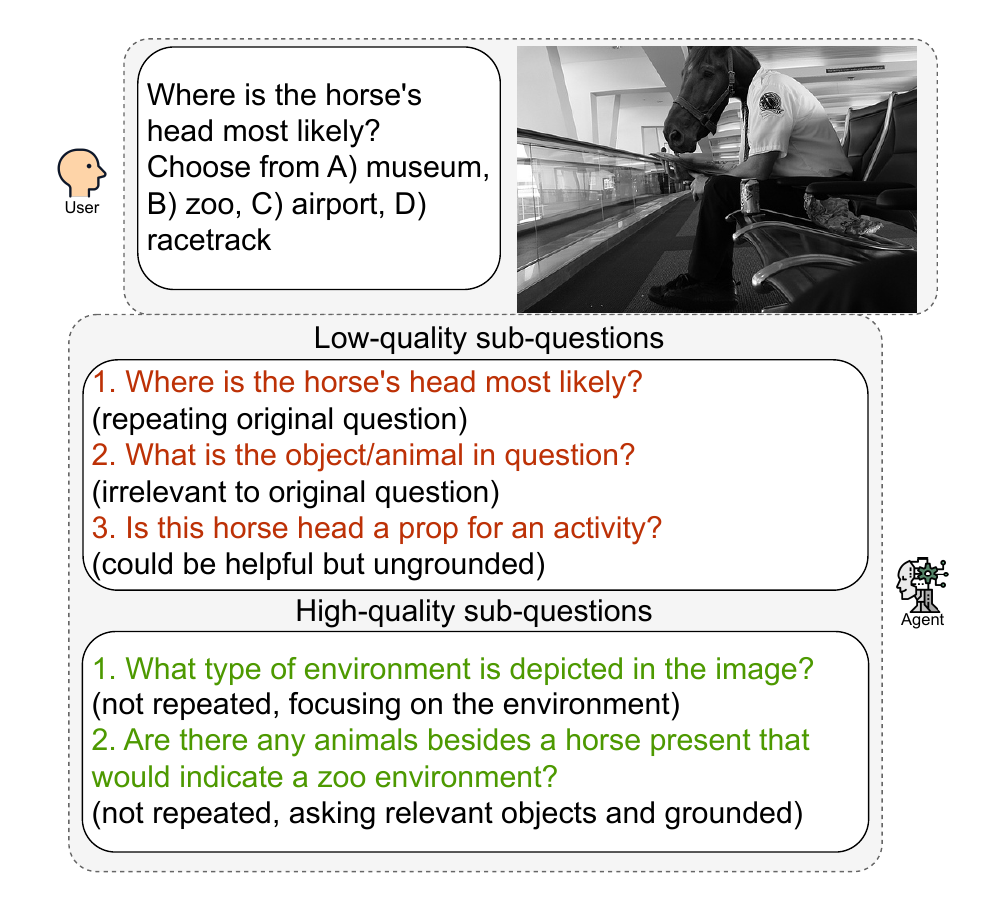}
\caption{Question decomposition examples of high quality and low quality given a certain image and question.}
\label{fig2}
\end{figure}

A more detailed explanation of the criteria with cases is given in Figure \ref{fig2}. By employing this evaluation framework, we have three quantifiable metrics for each sub-question. Algorithm \ref{algorithm_eva} visually demonstrates the complete evaluation process for each sub-question within this framework.

Moreover, we have constructed an evaluation benchmark dataset, since there is currently no dataset composed entirely of complex questions that require decomposition to answer, we construct an evaluation dataset. We manually selected 100 complex questions each from A-OKVQA~\citep{schwenk2022aokvqa} and VQA-Introspect~\citep{selvaraju2020squinting}, making a total of 200 questions worth decomposing. A-OKVQA serves as a benchmark necessitating a substantial understanding of external knowledge to formulate accurate responses. VQA-Introspect is a VQA dataset that contains a large number of samples that need complex visual reasoning to answer. A-OKVQA samples are in the form of multiple choice while VQA-Introspect provides open-ended questions. Since these two public datasets have a large number of simple questions which do not need decomposition to answer, we construct an evaluation dataset based on them instead of directly evaluating on them.
\input{figures/basic_performance.tex}
After establishing the evaluation framework, we choose GPT-4V as the scoring model due to its powerful comprehensive reasoning performance. To ensure that GPT-4V's judgments align with human judgments, we have conducted alignment experiments. As shown in Appendix~\ref{appendix:alignment}, the results demonstrate that the scoring gap between the judgments of GPT-4V and human beings is small. It is reliable to adopt GPT-4V as the scoring model.

We have measured the VQD ability of popular existing MLLMs with SubQuestRater, including MiniGPT-v2~\citep{chen2023minigptv2}, LLaVA-1.5~\citep{liu2024improved},  Qwen-VL-Chat~\citep{bai2023qwenvl} and InternVL-Chat-V1-5~\citep{chen2024far}. The results in Table~\ref{allmodels} show that these existing models cannot generate satisfactory sub-questions. 

%% file: figures/basic_performance.tex
\begin{table*}[!htb]
    \centering
    \resizebox{\linewidth}{!}{
    \begin{tabular}{lccccc}
        \toprule
        \textbf{Criteria} & \textbf{MiniGPT-v2} & \textbf{LLaVA-1.5} & \textbf{Qwen-VL-Chat} & \textbf{InternVL-Chat-V1-5} & \textbf{GPT-4V} \\
        \midrule
        \textbf{Non-Repetition} &47.52 &42.19 &32.10 & 82.41&\textbf{97.40} \\
        \textbf{Relevance} &36.65 & 37.33&27.15 &73.42 & \textbf{75.36}\\
        \textbf{Groundedness} &43.30 &44.17 & 26.49&78.01 & \textbf{84.57}\\
        \bottomrule
    \end{tabular}
    }
    
    \caption{Average scores of VQD ability on three criteria of popular existing MLLMs, evaluated with SubQuestRater. The performance of GPT-4V is also provided for reference.
    }
    \label{allmodels}
\end{table*}

%% file: chapters/4-enhancing_VQD_ability.tex
\section{Enhancing MLLM's Visual Question Decomposition Capability}

Given that the existing MLLMs have poor performance on VQD, this section further explores how to improve the VQD ability of MLLMs. An intuitive method to enhance the decomposition performance of MLLMs is to finetune the models on a dataset tailored for VQD. Specifically, we need a dataset to finetune the models, which exclusively focuses on complex questions with high-quality sub-questions in the view of Non-Repetition, Relevance, and Groundedness. However, there does not exist such a public VQA dataset. Therefore, we propose a specialized dataset, termed DecoVQA, to improve the VQD ability. Furthermore, for effective VQD, models also need to have an improved ability to decide when to decompose questions. We also explain the finetuning pipeline with our proposed dataset and a novel training objective to achieve that goal in detail, as discussed below.
\subsection{Dataset Construction of DecoVQA}
\paragraph{Question Selection \& Decomposition Annotation}
In our exploration of question decomposition for VQA in Table~\ref{allmodels}, we recognize that not all questions in existing benchmark datasets necessitate decomposition for answering. Many questions are straightforward and can be addressed without employing a decomposition strategy. Our focus, therefore, is on questions that demand complex reasoning, making them suitable candidates for the decomposition annotation. For this purpose, we have selected A-OKVQA and VQA-Introspect as our primary data source, as these two datasets contain complex questions requiring external knowledge and visual reasoning to answer respectively.

To identify appropriate samples from A-OKVQA and VQA-Introspect, we adopt specific pre-selection strategies, shown in Appendix ~\ref{preselection} detailedly. After that, we conduct a manual review and pick 200 complex samples that require decomposition from pre-selected samples. Then we manually annotated these samples with logical sub-questions. The details of the annotation process are shown in Appendix ~\ref{annotation}.



\paragraph{Dataset Statistics}
\label{dataset construction}
After decomposition annotation, we collected 100 samples from A-OKVQA and 100 samples from VQA-Introspect with high-quality sub-questions from the perspective of Non-Repetition, Relevance, and Groundedness. To prevent overfitting and catastrophic forgetting, we manually picked out another 100 samples from A-OKVQA and 100 samples from VQA-Introspect, which are simple and VQD doesn't contribute to higher performance for them. These simple samples are added to DecoVQA in the form of direct answering. Overall, DecoVQA has 400 balanced samples in total.

\begin{figure*}[!htb]
    \centering
    \begin{subfigure}[t]{0.32\textwidth}
        \includegraphics[width=\linewidth]{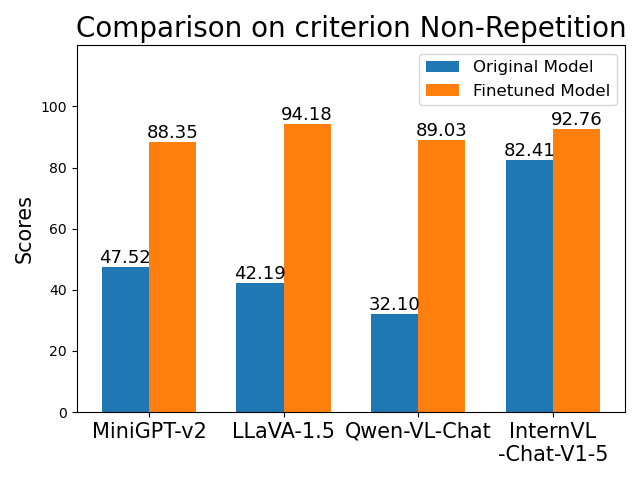}
    \end{subfigure}
    \hfill
    \begin{subfigure}[t]{0.32\textwidth}
        \includegraphics[width=\linewidth]{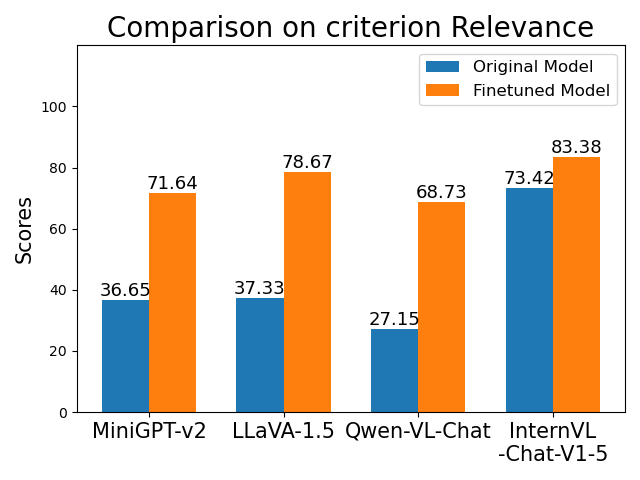}
    \end{subfigure}
    \hfill
    \begin{subfigure}[t]{0.32\textwidth}
        \includegraphics[width=\linewidth]{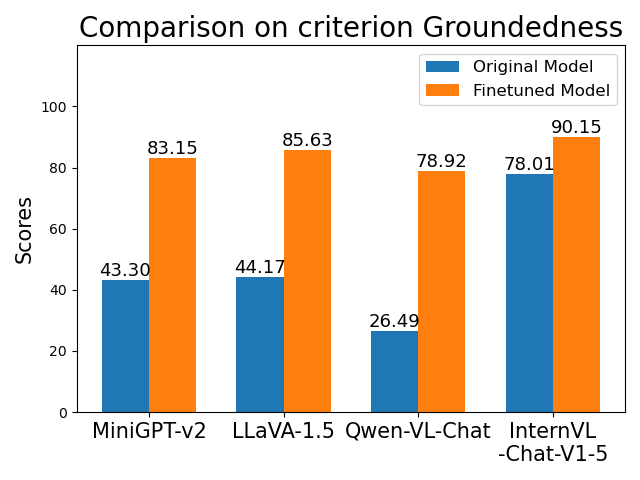}
    \end{subfigure}
    \caption{Comparison of VQD ability of different models across three evaluation criteria. Each bar chart represents a specific criterion, comparing the average scores of the original model (in blue) and the corresponding model finetuned with DecoVQA+ (in orange). The vertical axis shows the average scores, while the horizontal axis lists the models. The difference in bar height indicates the performance gain achieved through finetuning.}
    \label{fig:performance_gain}
\end{figure*}

\subsection{DecoVQA+}
To enhance the capability of MLLMs in selective decomposition, we add an extra QA round on the basis of DecoVQA to enable the models to learn when to decompose properly, facing questions with various difficulties. This extra QA round contains a query asking the models if they would directly answer without any decomposition, given an image and a question. The labels for simple questions are "yes" while the ones for complex questions with human-annotated sub-questions are "no". The extra QA round is added in front of all existing QA rounds of DecoVQA. We demonstrate the full prompt of a training sample in Figure~\ref{fig:prompts}. We refer to this upgraded version as "DecoVQA+".

The superiority of DecoVQA+ is proven in Appendix~\ref{appendix:dataset_comparison}, compared to the existing dataset VQA-Introspect. In the ablation study on DecoVQA+, as shown in Table~\ref{tab:ablation_ar_loss}, we compare the complete DecoVQA+ to the version with only 100 or 200 samples. The model achieves very similar results after being finetuned with different versions of DecoVQA+. On the one hand, the ablation study proves that our proposed dataset has sufficient samples to train the model. On the other hand, it also indicates that our finetuning pipeline remains efficient, even if there is a lack of high-quality finetuning data in most real-world cases.

To find out whether MLLMs learn to identify questions that need decomposition, we develop an evaluation dataset, Whether2Deco. It consists of 200 simple questions where direct answering is sufficient to answer them correctly and 200 complex questions that need VQD to answer, which are organized in the form of the extra round in DecoVQA+. The questions are equally sampled from A-OKVQA and VQA-Introspect. The statistics of all utilized public datasets and newly proposed datasets are shown in Table~\ref{tab:proposed_dataset}.


\subsection{Training Objective}
It is intuitive to finetune MLLMs to improve models' performance on VQD. However, directly applying the conventional next-token prediction loss (NTP loss) on the finetuning for \textit{selective} VQD may not be appropriate. To improve MLLMs' capability of identifying the questions that need to be decomposed, we propose a training objective, SelectiveVQD Loss, combining the NTP loss and a binary cross entropy loss (BCE loss). The BCE loss aims to penalize the errors in deciding whether to decompose, compared to the labels given in each sample of DecoVQA+. 

When the model is asked whether it would perform VQD, we firstly find the specific token position for "yes" or "no" in the sentence, select the logits of these two specific tokens in that position, i.e. "yes" and "no", and then transform these two logits into probabilities through softmax:
\begin{equation}
    \resizebox{\linewidth}{!}{
    $ 
    \begin{split}
        \mathbb{P}(yes)&=
        \mathbb{P}(\hat{w_s}="yes"|\hat{w_s}\in\{"yes", "no"\}) \\
        &= Softmax(logit(\hat{w_s}="yes"), logit(\hat{w_s}="no")) \\
        \mathbb{P}(no)&=
        \mathbb{P}(\hat{w_s}="no"|\hat{w_s}\in\{"yes", "no"\}) \\
        &= 1 - \mathbb{P}(yes), \\
    \end{split}
    $
}
\end{equation}
where $s$ is the specific token position in the sentence for "yes" or "no" and $w_s$ is the specific token of "yes" or "no".

We compute the BCE loss between these two probabilities, and compute the cumulative NTP loss across all conversation rounds for each sample:
\begin{equation}
    \resizebox{\linewidth}{!}{
    $
    BCELoss = -[y_slog\mathbb{P}(yes) + (1-y_s)log(1-\mathbb{P}(yes))]
    $
    }
\end{equation}
\begin{equation}
    \resizebox{\linewidth}{!}{
    $
    NTPLoss = -\sum_{i=1}^Mlog\mathbb{P}(\hat{w_i}=w_i|\hat{w_{i-1}},...,\hat{w_{1}}),
    $
    }
\end{equation}
where $y_s$ is the binary label indicating whether a specific sample needs decomposition or not. $w_i$ denotes the $i$-th token in the ground truth sentence while $\hat{w_i}$ denotes the predicted $i$-th token and $M$ is the number of tokens of the prediction. 

The final combined SelectiveVQD Loss is a weighted sum of both NTP loss and BCE loss:
\begin{equation}
    \resizebox{\linewidth}{!}{
    $
    SelectiveVQDLoss = \sum_{j=1}^N( \lambda \cdot NTPLoss_{j} +  \omega \cdot BCELoss_{j}),
    $
    \label{BCL}
    }
\end{equation}
where $j$ denotes the $j$-th training sample and $N$ denotes the total number of training samples. The NTP loss is computed for the entire training sample, and the BCE loss focuses specifically on determining whether to decompose the given question in the selective stage. $\lambda$ and $\omega$ are two tunable hyperparameters to balance the weights of the two losses in the final combined loss. 

%% file: chapters/5-experiments.tex
\section{Experiments} 
\subsection{Experiment Setup}
\input{tables/acc}
\paragraph{Models}
With \evalFrame, we compare the VQD ability of four popular MLLMs: MiniGPT-v2, LLaVA-1.5, Qwen-VL-Chat, and InternVL-Chat-V1-5 before and after finetuning on our proposed datasets. We finetune all these MLLMs on DecoVQA, DecoVQA+, and DecoVQA+ with SelectiveVQD Loss to see the improvement in their VQD capability. Additionally, we evaluate the improvement in VQA accuracy and the models' capability to appropriately determine when to decompose questions through finetuning.

\paragraph{Datasets} The finetuning datasets include DecoVQA and DecoVQA+. DecoVQA+ adds an extra QA round based on DecoVQA, asking models whether to decompose questions before decomposition. As for used evaluation datasets, we assess the VQD capability of MLLMs before and after finetuning on the proposed evaluation dataset in SubQuestRater, which contains 200 complex questions. The prompt for evaluating the VQD ability is shown in Figure~\ref{fig:prompts_subquestrater}. Besides, we evaluate the VQA accuracy on A-OKVQA, GQA~\citep{hudson2019gqa}, and VQA-Introspect, containing complex reasoning questions (please refer to Appendix~\ref{appendix:public_datasets} for more statistical details). As for A-OKVQA and VQA-Introspect, the subsets of data used for inference are different from the subsets selected for constructing our finetuning datasets, preventing the problem of data leakage. We also evaluate the accuracy on Whether2Deco to test whether MLLMs are able to determine when to decompose questions properly. The prompt for evaluation experiments on accuracy is under selective VQD setting, which is in the same form of samples in DecoVQA+ shown in Figure~\ref{fig:prompts}. 

\subsection{Quantitative Evaluation}
\label{quantitative}
The quantitative evaluation involves two parts: \textit{evaluation of decomposed sub-questions} under \evalFrame framework and \textit{accuracy comparison} on VQA datasets and Whether2Deco. The finetuning is efficient based on the dataset with a small number of samples. The supplementary evaluation on MMBench~\citep{liu2023mmbench} in Appendix~\ref{appendix:mmbench} shows our finetuning does not hurt the all-around performance of MLLMs, but even slightly improves the comprehensive performance in many aspects.

\paragraph{Evaluation of Decomposed Sub-questions}
To compare the VQD ability of the MLLMs before and after finetuning, we conduct evaluation with the SubQuestRater framework.
Figure~\ref{fig:performance_gain} illustrates the comparison of average scores of sub-questions generated by four MLLMs before and after finetuning with DecoVQA+. It can be observed that finetuned models have outperformed their original versions on all three criteria, indicating the VQD ability has been enhanced significantly through finetuning. Some of the models even show nearly double the scores in some criteria after finetuning. In addition to the average score, we also compare the number of samples before and after finetuning, which achieve a high score (75-100) and a low score (0-25) on three criteria, as shown in Figure~\ref{fig:numbers_vqd}. The results show that there are considerably more high-scored samples and less low-scored samples through finetuning. The VQD abilities of other finetuned checkpoints are listed in Table~\ref{ablation_100200}. It shows that all finetuned versions of MLLMs have an improvement in VQD ability compared to the original model.

Additionally, we have conducted experiments by varying the number of samples in DecoVQA, which ultimately lead to similar results, as shown in Table~\ref{4models}, indicating that there is no need to add more samples to DecoVQA+ for further improvements.

\paragraph{VQA Accuracy \& Whether2Deco Accuracy}
\begin{figure*}[!htb] 
\centering 
\includegraphics[width=\linewidth]{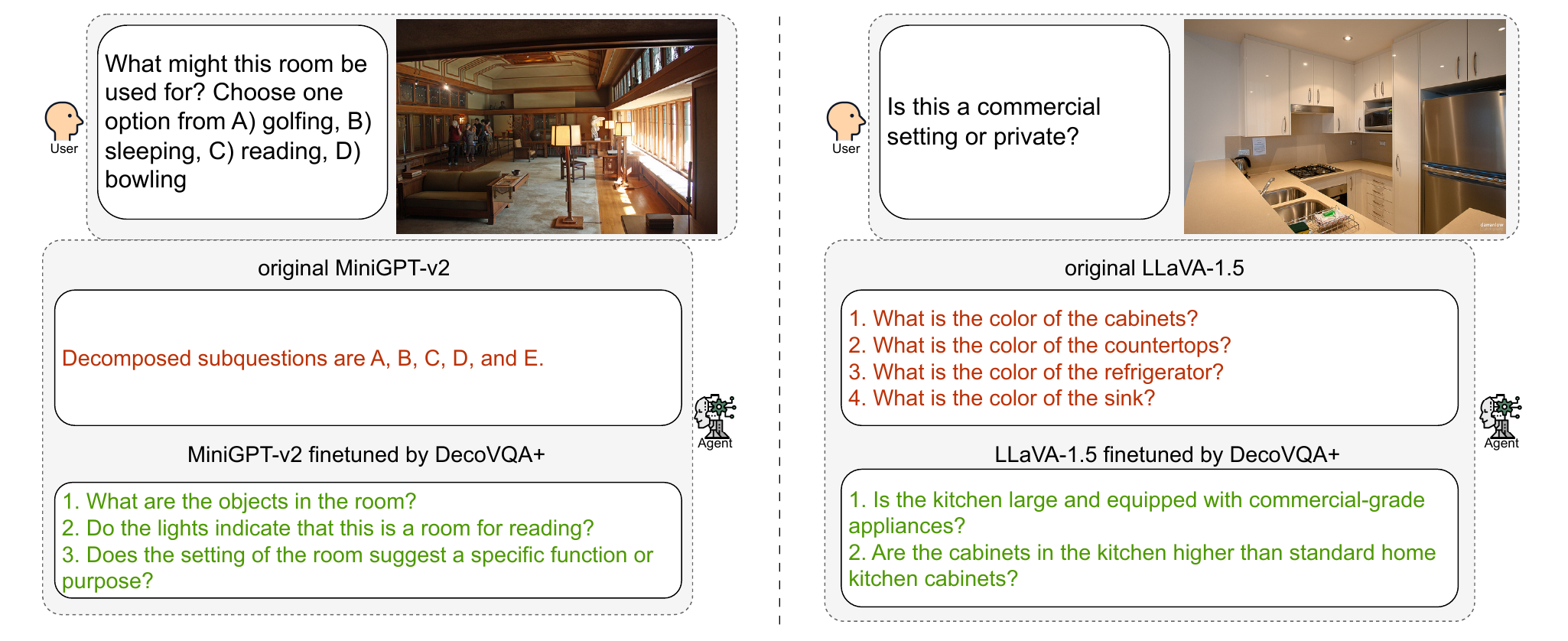}
\caption{Cases showing the comparison of question decomposition by different models before and after finetuning. The left image demonstrates MiniGPT-v2's decomposition on A-OKVQA, while the right image shows LLaVA-1.5's decomposition on VQA-Introspect.} 
\label{eg} 
\end{figure*}

Firstly, we investigate whether better VQD performance leads to higher accuracy. To prove that point, we compare the accuracy of models finetuned with DecoVQA and their corresponding baselines, as shown in the second and the first row of each model respectively in Table~\ref{tab:acc}. It is clear that the models with higher VQD capability through finetuning achieve higher accuracy than their original versions in most experiments.

Existing MLLMs are unable to decide when to decompose appropriately and tend to make a fifty-fifty guess, as shown in the first line of each model. For improving the performance in selective decomposition, it is very important for models to learn when to decompose, facing questions with varying difficulties, since unnecessary decomposition may mislead the reasoning process. From the second row of each model, we can see that higher quality of sub-questions does not mean better performance on determining when to decompose questions. To address this problem, we finetune the models with DecoVQA+. As shown in the third row of each model in Table~\ref{tab:acc}, there is a significant improvement in the accuracy on Whether2Deco after fine-tuning with DecoVQA+, compared to the baseline and the checkpoint finetuned by DecoVQA. Moreover, the accuracy of VQA tasks also increases because of the better whether-to-decompose policy of the finetuned models.

To further enhance the ability of MLLMs to perform selective decomposition, we train the models with SelectiveVQD Loss. In contrast to the training with only the NTP loss, the models achieve higher accuracy on Whether2Deco and also VQA tasks in most cases. If compared to the original models before finetuning, the accuracy on all evaluation datasets increases significantly. The results of evaluation experiments with different random seeds in Figure~\ref{fig:variance} show the stable effectiveness of our entire pipeline. We also compare our proposed VQD pipeline with the existing paradigm of unimodal QD based on the image caption, as shown in~\ref{tab:compare_language}. The results demonstrate that VQD outperforms the unimodal QD method. The comparison between our finetuning and In-context Learning proposed in~\citep{NEURIPS2023_b14cf0a0} is shown in Appendix~\ref{compare_ICL}.

\subsection{Qualitative Evaluation}
In this subsection, we will use several examples to visually illustrate the changes of sub-questions before and after finetuning with DecoVQA+. Figure~\ref{eg} shows that the quality of the sub-questions has indeed been significantly improved after finetuning. The sub-questions generated by finetuned models are not repetitive, relevant to the original question and grounded, instead of ineffective decomposition or low-quality sub-questions originally. More case studies are shown in Figure~\ref{fig:more_case_studies}.

%% file: tables/acc.tex
\begin{table*}[!htb]
    \centering
    \resizebox{\linewidth}{!}{
    \begin{tabular}{lllll}
        \toprule
        \textbf{Models} & \textbf{A-OKVQA} & \textbf{GQA} & \textbf{VQA-Introspect} & \textbf{Whether2Deco}\\
        \midrule
         
        \textbf{MiniGPT-v2} & 41.2 & 44.2 & 62.1 & 46.8\\
        \quad finetuned by DecoVQA & 60.6 \up{$\uparrow$ (+19.4)} & 50.4 \up{$\uparrow$ (+6.2)} & 71.8 \up{$\uparrow$ (+9.7)} & 42.8 \down{$\downarrow$ (-4.0)}\\
        \quad finetuned by DecoVQA+ & 60.7 \up{$\uparrow$ (+19.5)} & 50.7 \up{$\uparrow$ (+6.5)} & 72.1 \up{$\uparrow$ (+10.0)} & 61.0 \up{$\uparrow$ (+14.2)}\\
        \quad finetuned by DecoVQA+ with SelectiveVQD Loss & \textbf{64.0} \up{$\uparrow$ (+22.8)} & \textbf{51.7} \up{$\uparrow$ (+7.5)} & \textbf{72.5} \up{$\uparrow$ (+10.4)} & \textbf{71.5} \up{$\uparrow$ (+24.7)}\\
        \noalign{\vskip 1ex}\cdashline{1-5}\noalign{\vskip 1ex}
        
        \textbf{LLaVA-1.5} & 67.7 & 52.1 & 67.2 & 49.3\\
        \quad finetuned by DecoVQA & 69.4 \up{$\uparrow$ (+1.7)} & 52.8 \up{$\uparrow$ (+0.7)} & 73.5 \up{$\uparrow$ (+6.3)} & 4.8* \down{$\downarrow$ (-44.5)}\\
        \quad finetuned by DecoVQA+ & 72.7 \up{$\uparrow$ (+5.0)} & \textbf{57.2} \up{$\uparrow$ (+5.1)} & 75.4 \up{$\uparrow$ (+8.2)} & 68.8 \up{$\uparrow$ (+19.5)}\\
        \quad finetuned by DecoVQA+ with SelectiveVQD Loss & \textbf{73.9} \up{$\uparrow$ (+6.2)} & 56.7 \up{$\uparrow$ (+4.6)} & \textbf{75.8} \up{$\uparrow$ (+8.6)} & \textbf{75.0} \up{$\uparrow$ (+25.7)}\\

        \noalign{\vskip 1ex}\cdashline{1-5}\noalign{\vskip 1ex}

        \textbf{Qwen-VL-Chat} & 71.4 & 53.5 & 77.8 & 48.0\\
        \quad finetuned by DecoVQA & 72.0 \up{$\uparrow$ (+0.6)} & 58.0 \up{$\uparrow$ (+4.5)} & 75.9 \down{$\downarrow$ (-1.9)} & 43.3 \down{$\downarrow$ (-4.7)}\\
        \quad finetuned by DecoVQA+ & 73.1 \up{$\uparrow$ (+1.7)} & \textbf{59.3} \up{$\uparrow$ (+5.8)} & 83.6 \up{$\uparrow$ (+5.8)} & 58.8 \up{$\uparrow$ (+10.8)}\\
        \quad finetuned by DecoVQA+ with SelectiveVQD Loss & \textbf{73.3} \up{$\uparrow$ (+1.9)} & 59.1 \up{$\uparrow$ (+5.6)} & \textbf{83.9} \up{$\uparrow$ (+6.1)} & \textbf{61.8} \up{$\uparrow$ (+13.8)}\\

        \noalign{\vskip 1ex}\cdashline{1-5}\noalign{\vskip 1ex}

        \textbf{InternVL-Chat-V1-5} & 80.7 & 64.8 & 80.5 & 58.3\\
        \quad finetuned by DecoVQA & 83.5 \up{$\uparrow$ (+2.8)} & 66.4 \up{$\uparrow$ (+1.6)} & 86.0 \up{$\uparrow$ (+5.5)} & 53.5 \down{$\downarrow$ (-4.8)}\\
        \quad finetuned by DecoVQA+ & 83.3 \up{$\uparrow$ (+2.6)} & 66.5 \up{$\uparrow$ (+1.7)} & 86.9 \up{$\uparrow$ (+6.4)} & 67.0 \up{$\uparrow$ (+8.7)}\\
        \quad finetuned by DecoVQA+ with SelectiveVQD Loss & \textbf{83.7} \up{$\uparrow$ (+3.0)} & \textbf{66.8} \up{$\uparrow$ (+2.0)} & \textbf{87.3} \up{$\uparrow$ (+6.8)} & \textbf{68.3} \up{$\uparrow$ (+10.0)}\\

        \bottomrule
    \end{tabular}
    }
    \caption{Comparison of VQA accuracy (\%) on external knowledge (A-OKVQA) and visual reasoning (GQA and VQA-Introspect) datasets and Whether2Deco accuracy (\%) before and after fine-tuning MLLMs. DecoVQA+ is constructed based on DecoVQA, with an extra QA round asking MLLMs whether the question needs VQD to answer or not. *Here LLaVA-1.5 fails to follow the pre-defined answering template.
    }
    \label{tab:acc}
\end{table*}

%% file: chapters/6-conclusion.tex
\section{Conclusion}

This paper systematically investigates VQD capabilities on MLLMs. We propose a systematic evaluation framework for VQD, SubQuestRater, including a dataset and evaluation metrics to quantitatively measure the generated sub-questions by MLLMs. SubQuestRater is applied to popular MLLMs and we find that they are inadequate to produce high-quality sub-questions. To enhance the capability of MLLMs to decompose questions, a specialized dataset DecoVQA with human-annotated sub-questions is proposed. To further improve the ability to perform selective VQD, we propose a training pipeline with an upgraded dataset DecoVQA+ and a novel training objective. Finetuned MLLMs demonstrate significant improvement in the quality of generated sub-questions and the policy of whether-to-decompose. Additionally, the models also achieve higher VQA accuracy under selective VQD through finetuning on our proposed datasets.

%% file: chapters/limitation.tex
\section*{Limitations}
The main limitations in our work include: 1) Question Decomposition can be extended into complex task decomposition for an agent (multiple sub-tasks), leaving it as future work. 2) We apply finetuning to increase MLLM's VQD ability, which requires the model's detailed parameter information. Thus, community users could not apply our method for enhancing closed-source MLLMs.

%% file: chapters/appendix.tex
\appendix
\section{Alignment of Judgements from GPT-4V and Human Reviewers}
To ensure that the judgments from GPT-4V and human beings are highly aligned, human reviewers manually evaluate the sub-questions generated by MiniGPT-v2 and its finetuned checkpoint on three criteria defined in SubQuestRater. We compare the judgments from GPT-4V and human reviewers from three perspectives: 1) Comparison of average score, 2) Pearson correlation coefficient~\citep{freedman2007statistics}, 3) Spearman correlation coefficient~\citep{zar2005spearman}. The comparison of average score serves as a coarse-grained evaluation of overall alignment, with results shown in Table~\ref{tab:alignment_average_score}. Pearson and Spearman correlation coefficients assess linear and monotonic relationships between two sets of judgments respectively, with results shown in Table~\ref{tab:alignment_pearson_spearman}. The results demonstrate that the judgments from GPT-4V and human reviewers on all three criteria are highly aligned.
\label{appendix:alignment}
\input{tables/alignment_average_score}
\input{tables/alignment_pearson_spearman}
\section{Ablation Studies}
\label{appendix:ablation}

The results of ablation studies, as shown in Table~\ref{4models} and in Table~\ref{tab:ablation_ar_loss}.  The samples from all the versions of DecoVQA+ follow a balanced distribution. For instance, DecoVQA+100 has 25 complex MC and 25 complex open-ended questions that need VQD while it has the other 25 simple MC and 25 simple open-ended questions that do not need VQD to answer. The models finetuned by DecoVQA+ with varying sample numbers generate similar results on VQD ability, VQA accuracy, and Whether2Deco accuracy, demonstrating that 400 samples are sufficient to ensure both efficient and reliable finetuning.
\input{figures/4modelsperformance.tex}
\input{tables/ablation_50_100_200}

\section{Experiment Details}
\label{appendix:experiments}

\subsection{Models}
The versions of models we use are listed as follows, corresponding official tokenizers are applied for all the models:
\begin{itemize} [leftmargin=*]
\itemsep0em
\item\textbf{MiniGPT-v2}, which is based on Llama2-Chat-7B-HF~\citep{touvron2023llama}.
\item\textbf{LLaVA-1.5}, which is based on Vicuna-13B v1.5~\citep{vicuna2023} and with lora as the pretraining schedule.
\item\textbf{Qwen-VL-Chat}, which is based on Qwen-7B~\citep{bai2023qwen}.
\item\textbf{InternVL-Chat-V1-5}, which is based on InternLM2-20B~\citep{cai2024internlm2}.
\item\textbf{GPT-4-vision-preview}, which is based on GPT-4~\citep{openai2024gpt4}.
\end{itemize}

\subsection{Finetuning Settings}
For all the mentioned open-source MLLMs, we use their official GitHub repository code to perform LoRA finetuning on the connection layer between the two modalities. All of the models are trained on 2 $\times$ A40 GPU until the training loss converges.

\subsection{Inference Settings}
We use a batch size of 1 in all inference tasks. Greedy search is used for all inferences. The parameters may be sub-optimal.

\subsection{Prompts}
\label{prompt}
\subsubsection{Prompt for Scoring VQD Ability}
\begin{figure*}[htbp]
    \centering
    \begin{subfigure}[c]{\textwidth}
    \centering
        \includegraphics[width=0.75\linewidth]{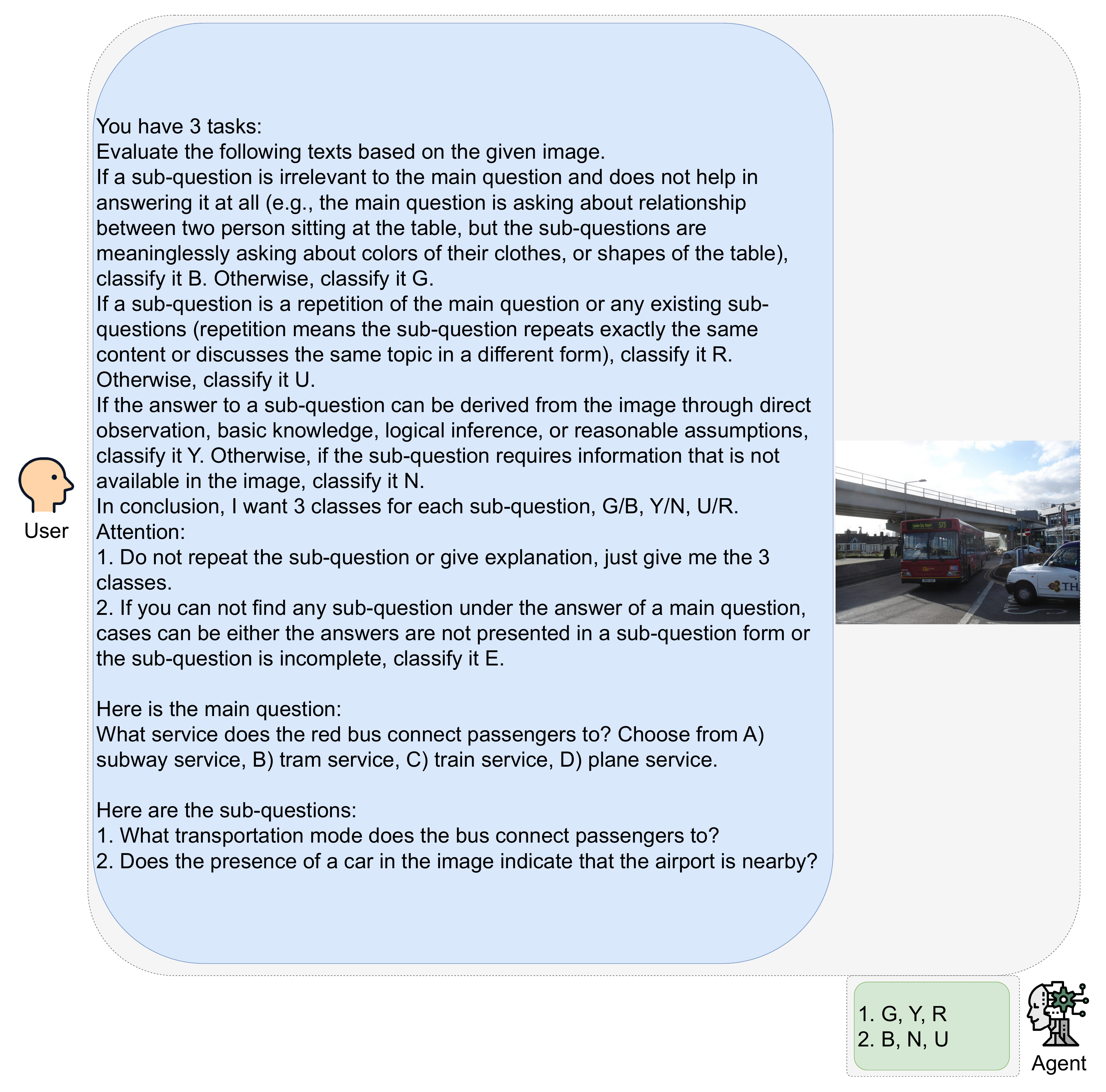}
        \caption{An example of evaluation on effective sub-questions.}
        \label{fig:prompt_subquestrater_effective}
    \end{subfigure}
    \hfill
    \begin{subfigure}[c]{\textwidth}
    \centering
        \includegraphics[width=0.75\linewidth]{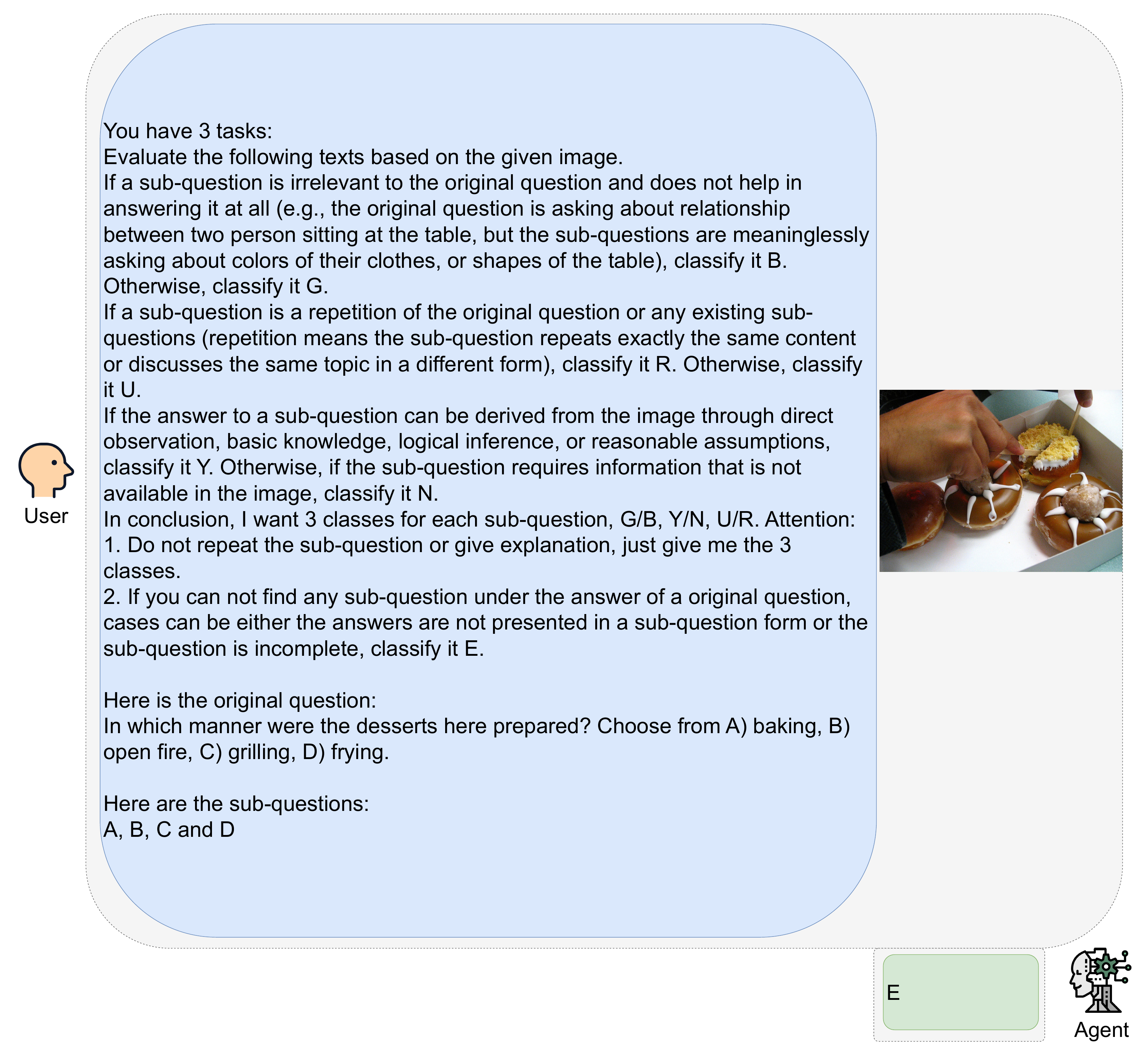}
        \caption{An example of evaluation on ineffective sub-questions (error).}
        \label{fig:prompt_subquestrater_error}
    \end{subfigure}
    \caption{Prompt for scoring the quality of sub-questions with GPT-4V.}
    \label{fig:prompts_subquestrater}
\end{figure*}

The complete prompt for scoring VQD ability, or specifically, the quality of sub-questions is shown in Figure~\ref{fig:prompts_subquestrater}.
\subsubsection{Prompt for Selective VQD}
\begin{figure*}[htbp]
    \centering
    \begin{subfigure}[c]{0.48\textwidth}
        \includegraphics[width=\linewidth]{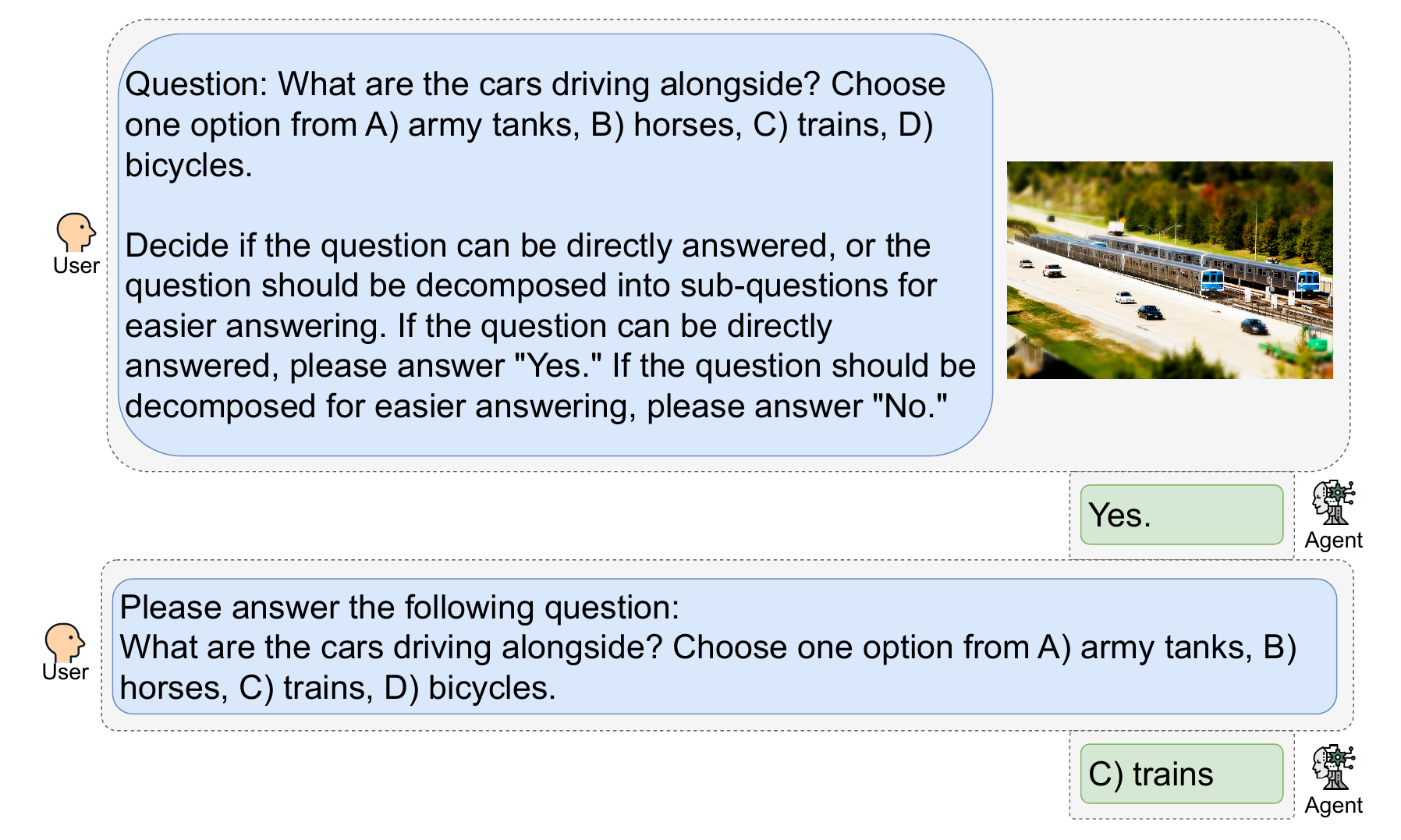}
        \caption{An example of prompt when the model chooses to directly answer the given question.}
        \label{fig:prompt_direct}
    \end{subfigure}
    \hfill
    \begin{subfigure}[c]{0.48\textwidth}
        \includegraphics[width=\linewidth]{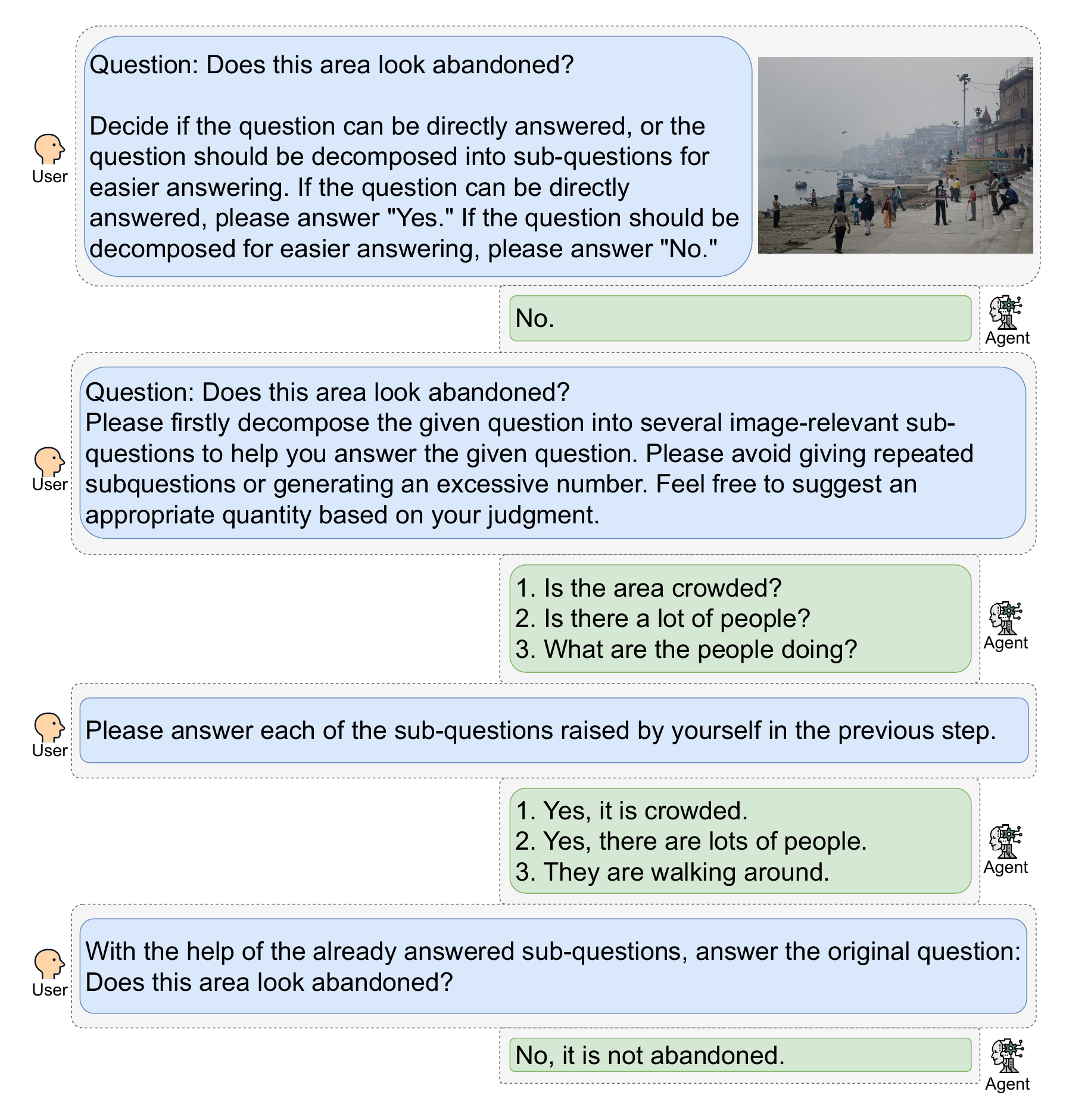}
        \caption{An example of prompt when the model chooses to decompose the given question.}
        \label{fig:prompt_decompose}
    \end{subfigure}
    \caption{Prompt of selective decomposition samples in DecoVQA+.}
    \label{fig:prompts}
\end{figure*}

As shown in Figure~\ref{fig:prompts}, firstly, we perform a selective stage, which asks the model whether to decompose the question. If the model answers that it can directly answer the question without decomposition by "Yes",
then implement the direct answering; if the model answers that it needs to decompose the question firstly by "No", then implement a three-phase decomposition process.

\section{Datasets}
\label{appendix:datasets}
\subsection{Public Datasets}
\label{appendix:public_datasets}
The statistics of used public datasets are listed in Table~\ref{tab:dataset}
.
\begin{itemize} [leftmargin=*]
\itemsep0em
\item\textbf{A-OKVQA}~\citep{schwenk2022aokvqa} is a complex knowledge-based benchmark for VQA. As an augmented version of OK-VQA~\citep{marino2019okvqa}, the questions in A-OKVQA are not only diverse but also require a wide-ranging commonsense and knowledge outside the image to answer. A-OKVQA has both multiple choice and open-ended question forms for each sample, and we select multiple choice here to cover more question types in the experiments. 
\item\textbf{GQA}~\citep{hudson2019gqa} features compositional questions related to real-world images, utilizing semantic representations of both scenes and questions to reduce language priors and conditional influences.
\item\textbf{VQA-Introspect}~\citep{selvaraju2020squinting} is a new dataset based on a reasoning split from VQA~\citep{VQA} dataset, which contains complex reasoning questions with the open-ended form. VQA-Introspect consists of 200K perception questions as sub-questions to help answer difficult reasoning questions. Though this public dataset provides us a large number of sub-questions, questions with varying difficulties are mixed together and no label is pointing it out, which leads to bad finetuning results for selective decomposition. We randomly sampled 3,000 questions for the evaluation on VQA-Introspect.
\end{itemize}
\input{tables/dataset_overview}
\subsection{Proposed Datasets}
\input{tables/proposed_dataset_overview}
The statistics of proposed datasets are listed in Table~\ref{tab:proposed_dataset}.

\section{Quantitative Evaluation for VQD Ability}
\input{figures/criteria.tex}

\begin{figure}[htbp]
\centering
\includegraphics[width=\linewidth]{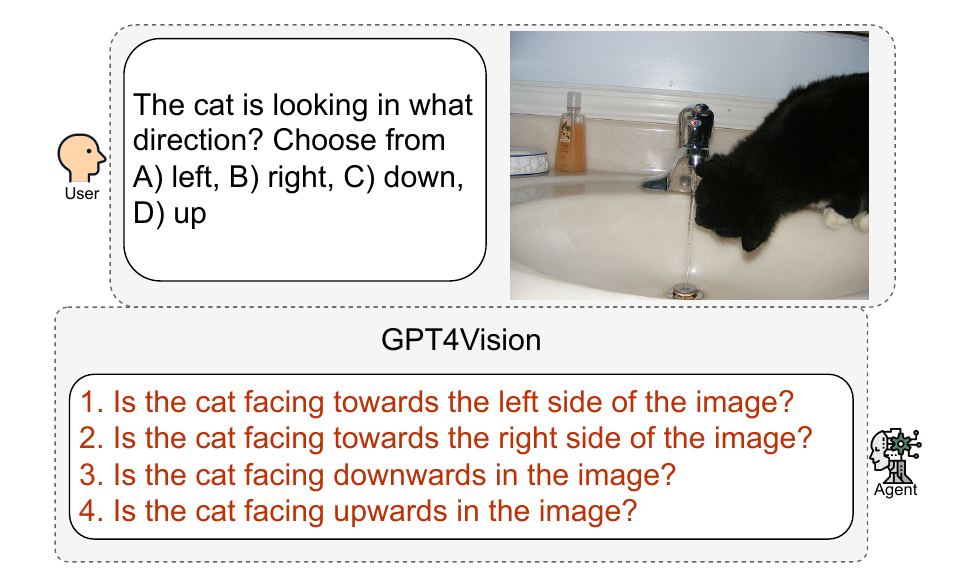}
\caption{In some cases, GPT-4V will also produce sub-questions that do not fit our criteria.}
\label{gptf}
\end{figure}

\begin{figure*}[htbp]
    \centering
    \begin{subfigure}[t]{0.32\textwidth}
        \includegraphics[width=\linewidth]{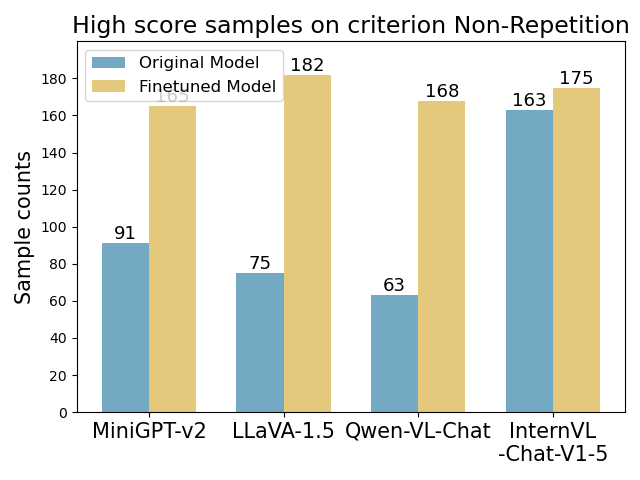}
    \end{subfigure}
    \hfill
    \begin{subfigure}[t]{0.32\textwidth}
        \includegraphics[width=\linewidth]{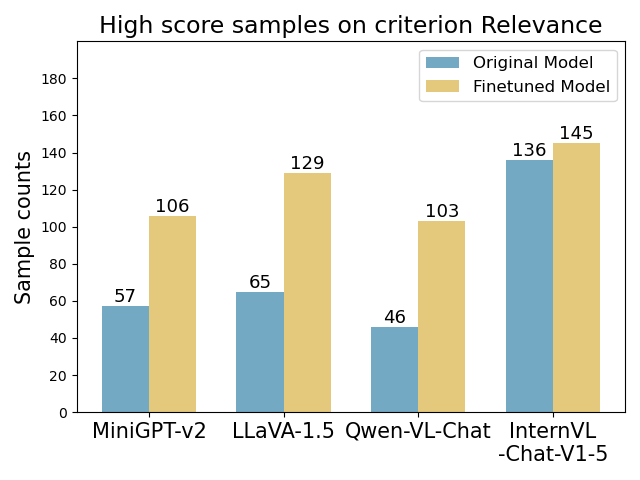}
    \end{subfigure}
    \hfill
    \begin{subfigure}[t]{0.32\textwidth}
        \includegraphics[width=\linewidth]{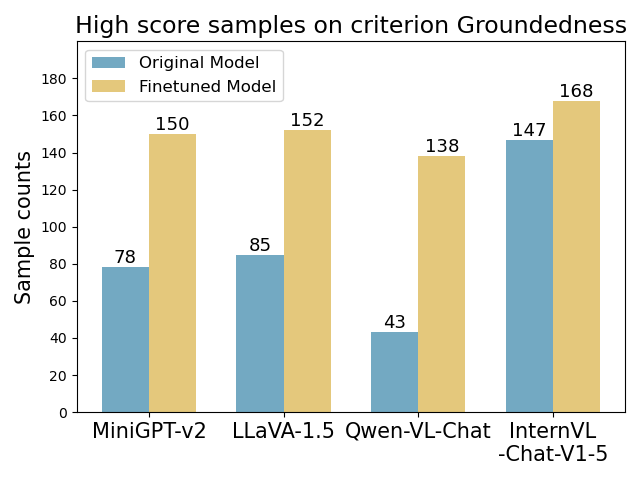}
    \end{subfigure}
    \hfill
    \begin{subfigure}[t]{0.32\textwidth}
        \includegraphics[width=\linewidth]{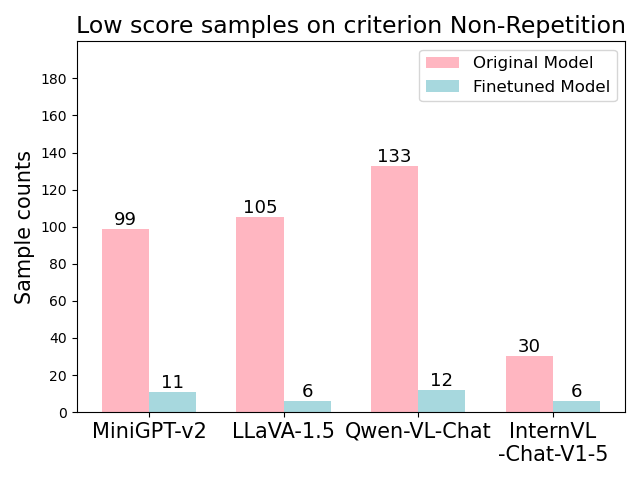}
    \end{subfigure}
    \hfill
    \begin{subfigure}[t]{0.32\textwidth}
        \includegraphics[width=\linewidth]{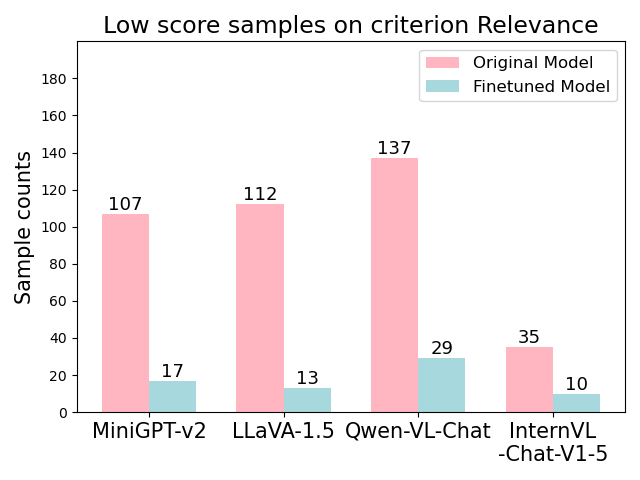}
    \end{subfigure}
    \hfill
    \begin{subfigure}[t]{0.32\textwidth}
        \includegraphics[width=\linewidth]{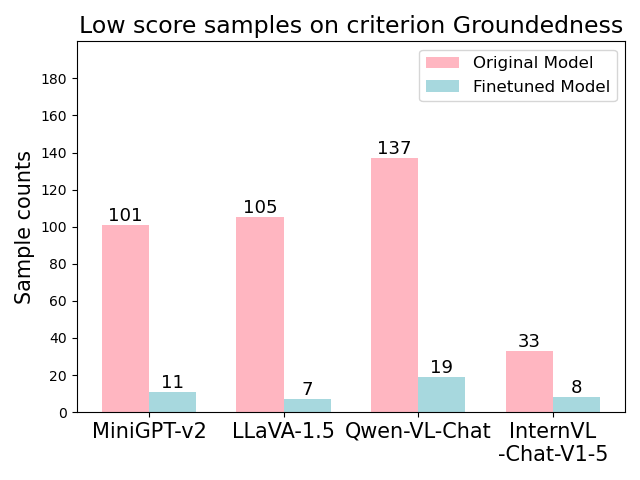}
    \end{subfigure}
    \caption{Comparison of VQD ability of different models across three evaluation criteria. Each bar chart represents a specific criterion. The first row compares the number of the high-scored (75-100) samples generated by the original model (in cyan) and the corresponding model finetuned with DecoVQA+ (in yellow). The second row compares the number of the low-scored (0-25) samples generated by the original model (in pink) and the corresponding model finetuned with DecoVQA+ (in blue). The vertical axis shows the number of high-scored samples or low-scored samples, while the horizontal axis lists the models. The difference in bar height indicates the performance gain achieved through finetuning.}
    \label{fig:numbers_vqd}
\end{figure*}

\input{figures/ablation.tex}

\section{Details of Data Construction}
\subsection{Pre-selection Strategies for Selecting Samples}
\label{preselection}
To identify questions from the A-OKVQA dataset that would benefit from decomposition, we employ a specific pre-selection strategy. Initially, we used a MLLM to perform zero-shot inference on the dataset, a process we term "direct inference". Subsequently, we engaged the same model in another round of zero-shot inference, but this time utilizing a question decomposition prompt. In this round, the model is asked to decompose the main question into sub-questions, then answer these, and finally proceed to answer the main question. We refer to this method as "decompose inference". We choose MiniGPT-v2 as the multimodal LLM here. Our primary focus was on questions that were incorrectly answered in direct inference but correctly in decompose-inference, as these exhibited a high likelihood of requiring decomposition. 

To find appropriate samples from the VQA-Introspect dataset, we adopt an automated pre-selection strategy based on BLEU~\citep{papineni-etal-2002-bleu} metric. BLEU metric is originally used to measure the quality of machine translation, here we use it as a metric to assess repetition. Since VQA-Introspect has provided a large number of redundant sub-questions, we firstly filter out semantically repetitive sub-questions for each sample to prevent from overfitting.  A higher BLEU score between two sub-questions means that one of the sub-questions is repetitive. Then we set a threshold number and choose the samples with a remaining number of sub-questions exceeding the threshold.

\subsection{Annotation Process}
\label{annotation}
Given the proficiency of GPT-4V in VQD, as shown in Table~\ref{allmodels}, we utilize GPT-4V to generate initial sets of decomposed sub-questions for each selected sample. Subsequently, we perform a meticulous manual review to these sub-questions. During this process, we eliminate sub-questions that do not contribute meaningfully towards answering the main question and also remove redundant sub-questions that share similar semantic content. Additionally, we supplement the sets with new sub-questions in instances where the decomposition logic appears incomplete, ensuring a more thorough and effective decomposition process.

\section{Robust Evaluation for MC Datasets}
To compute the accuracy of the inference results under multiple choice setting, since an exact match of either option index or word can lead to serious underestimation, the first step is to map the model answer into one of four options. We have designed a robust algorithm to evaluate the accuracy of multiple choice based on the method provided by A-OKVQA. As demonstrated in Alogrithm~\ref{algorithm_mc}, if no or several options are detected in the model answer during the exact match step, we use SentenceTransformer~\citep{reimers2019sentencebert} to map the model answer to one option or use GPT-4 to do the mapping when the answer is too long. We have observed that if the answer sentence is too long, especially when there is more than one option mentioned in the answer, the mapping by SentenceTransformer tends to be random and misleading.
For computing the accuracy over open-ended questions, given that the reference answer in used datasets has only one or two words, if the reference answer is mentioned in the model output, the output is considered as correct.
\input{figures/eval_mc}

\section{Variance}
\begin{figure}[!h]
    \centering
    \begin{subfigure}{\linewidth}
        \includegraphics[width=\linewidth]{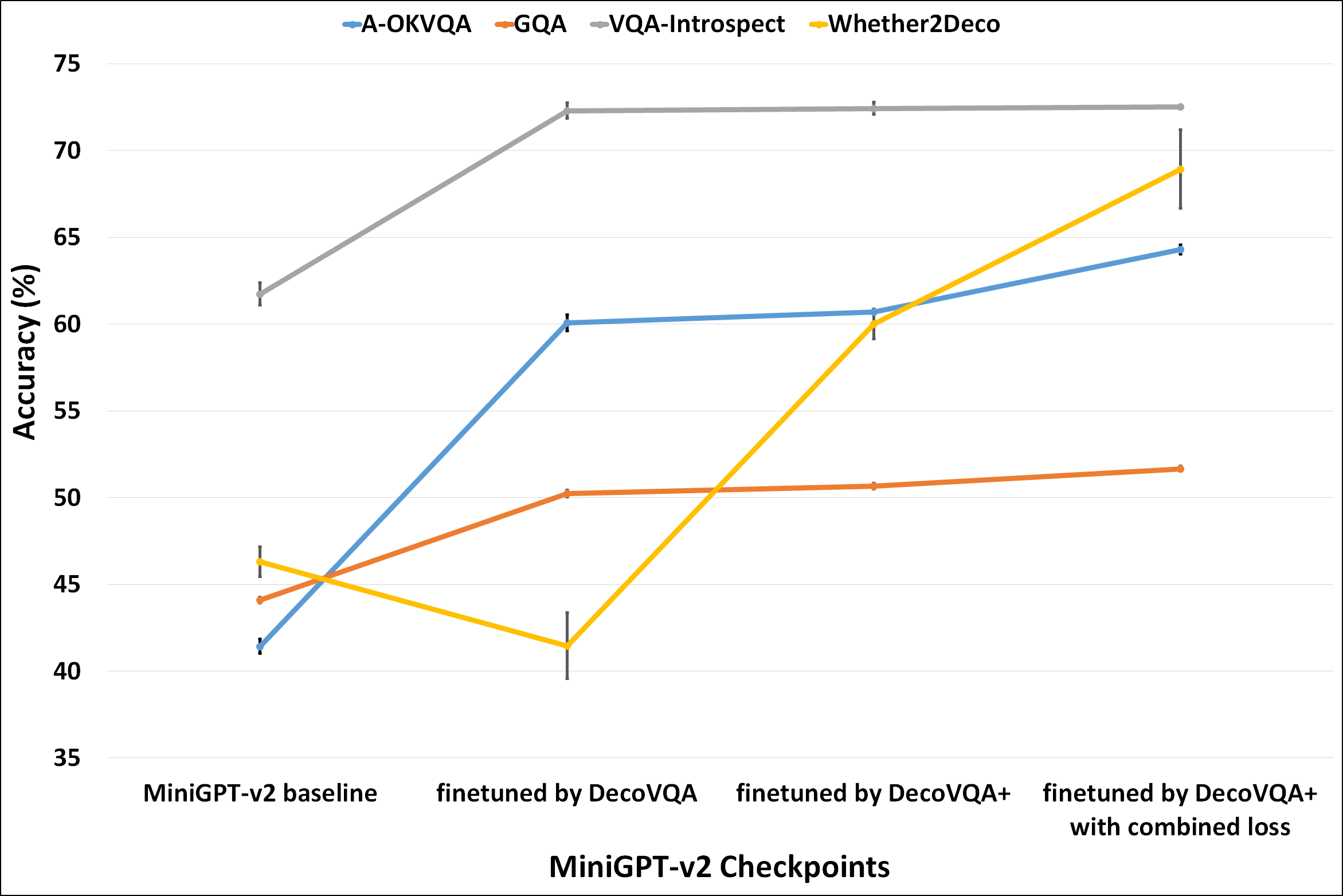}
        \vspace{0.6cm}
    \end{subfigure}
    
    \begin{subfigure}{\linewidth}
        \includegraphics[width=\linewidth]{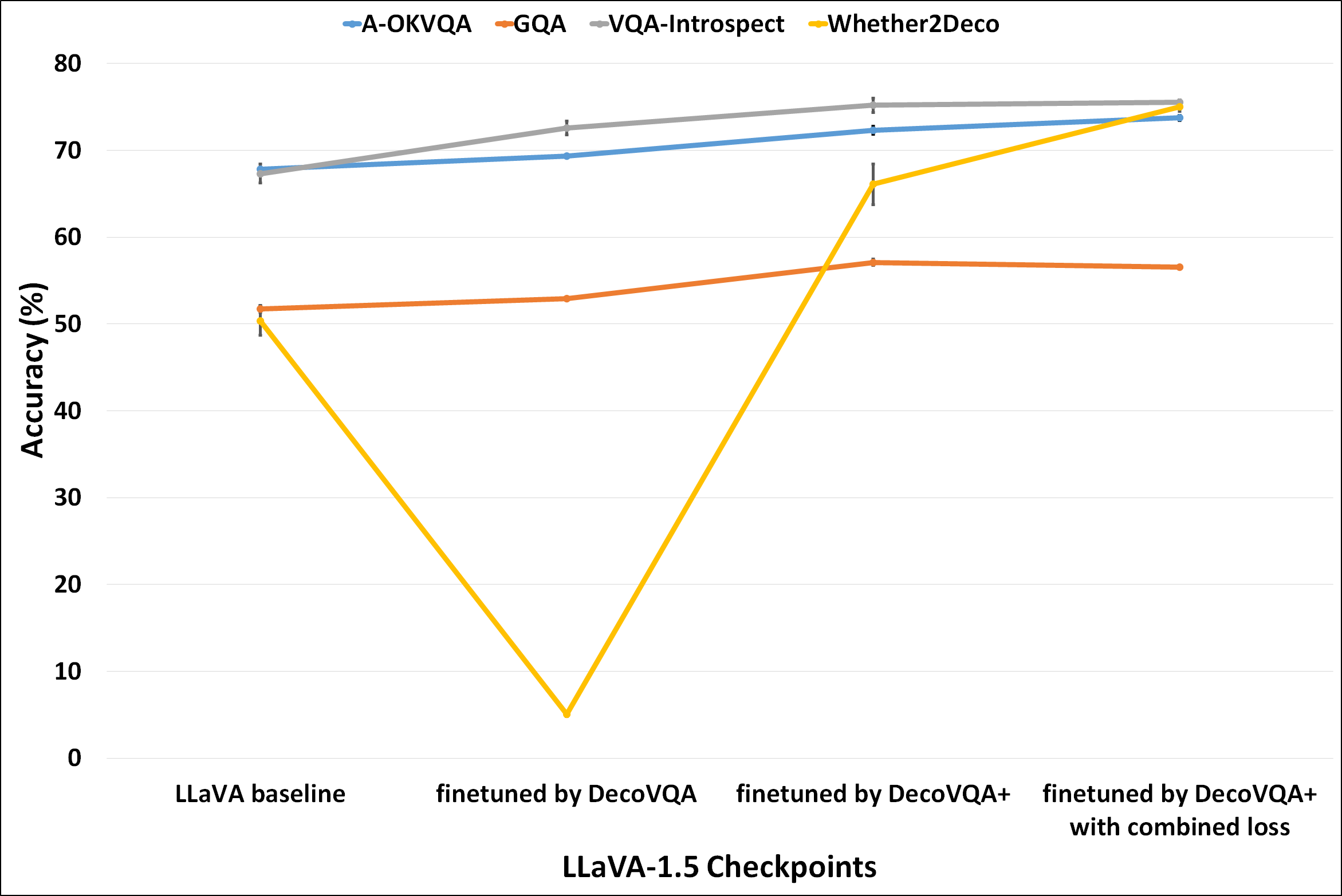}
    \end{subfigure}
\caption{Variance of inference experiments with MiniGPT-v2 and LLaVA-1.5, plotted as error bars. Each experiment is conducted with three different random seeds, keeping other settings unchanged.}
\label{fig:variance}
\end{figure}

To verify the stability of our proposed method, each experiment was done with three different random seeds, while keeping other settings unchanged. The variance results in Figure~\ref{fig:variance} show that random seeds influence the accuracy of the model output very slightly.

\section{Does the finetuning hurt the all-around performance?}
\label{appendix:mmbench}
\begin{figure*}[htbp]
\centering
    \includegraphics[width=\textwidth]{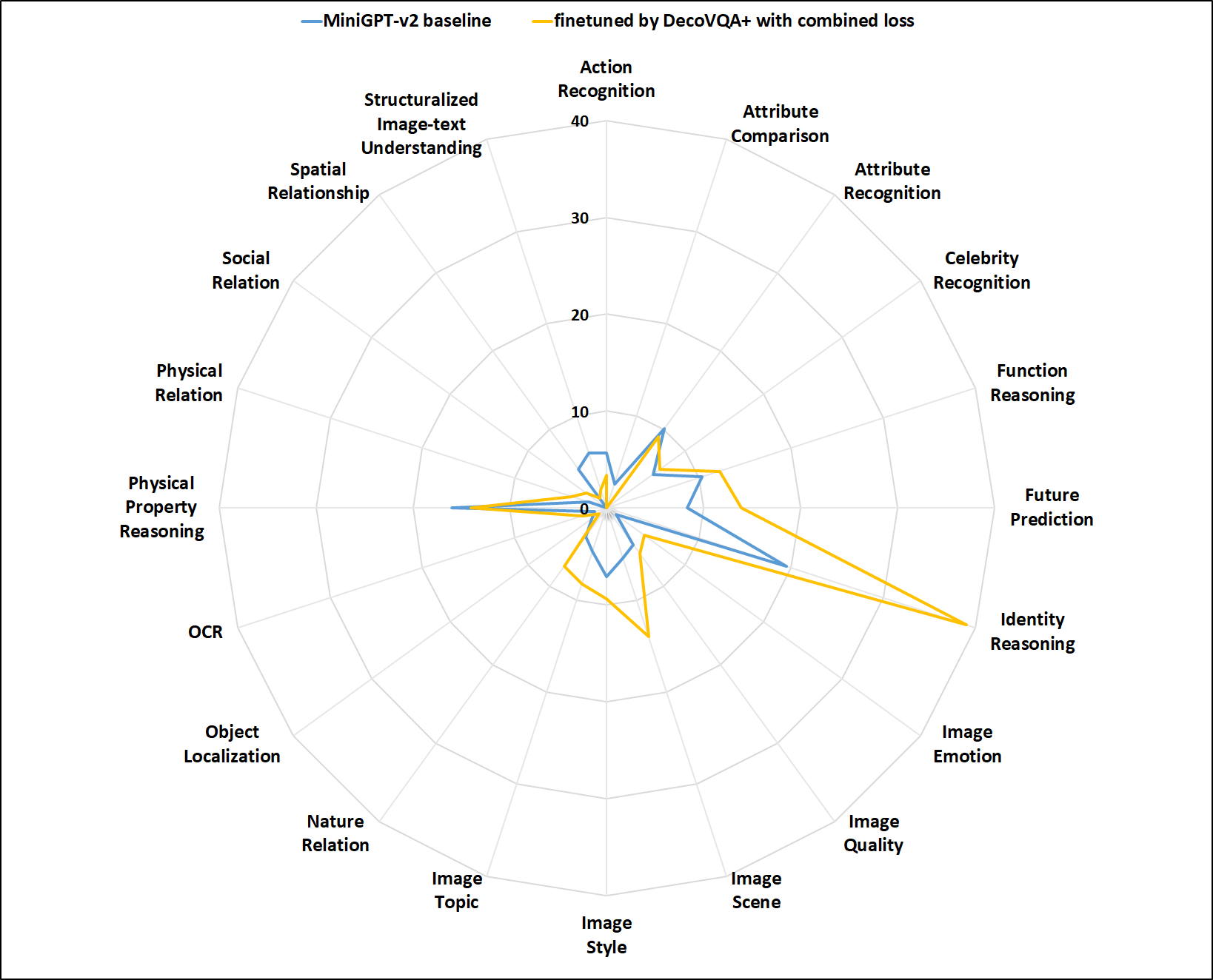}
\caption{Results of different checkpoints of MiniGPT-v2 across the 20 L-3 ability dimensions defined in MMBench.}
\label{fig:minigptv2_mmbench}
\end{figure*}

\begin{figure*}[htbp]
\centering
    \includegraphics[width=\textwidth]{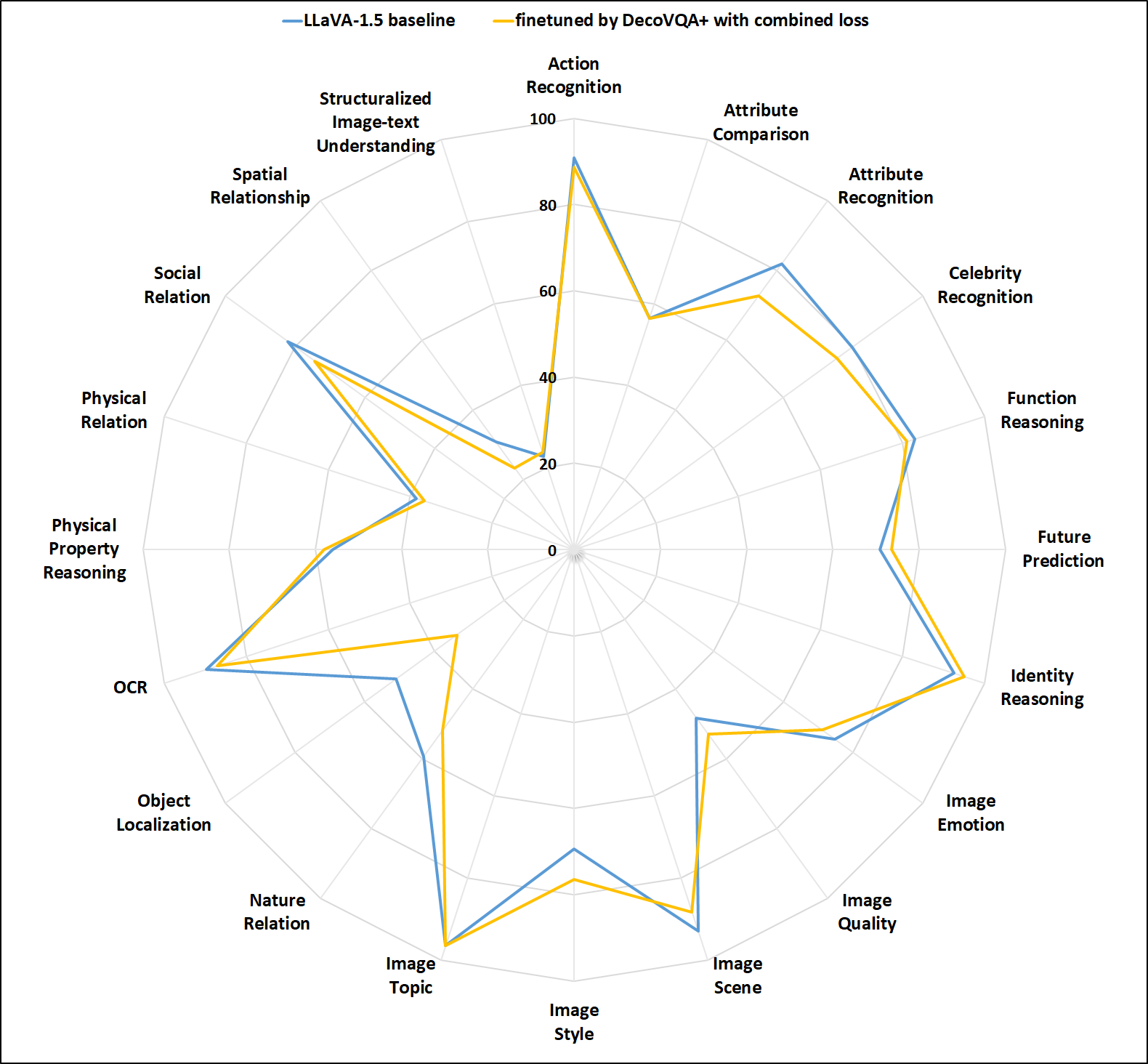}
\caption{Results of different checkpoints of LLaVA-1.5 across the 20 L-3 ability dimensions defined in MMBench.}
\label{fig:llava_mmbench}
\end{figure*}

\begin{figure*}[htbp]
\centering
    \includegraphics[width=\textwidth]{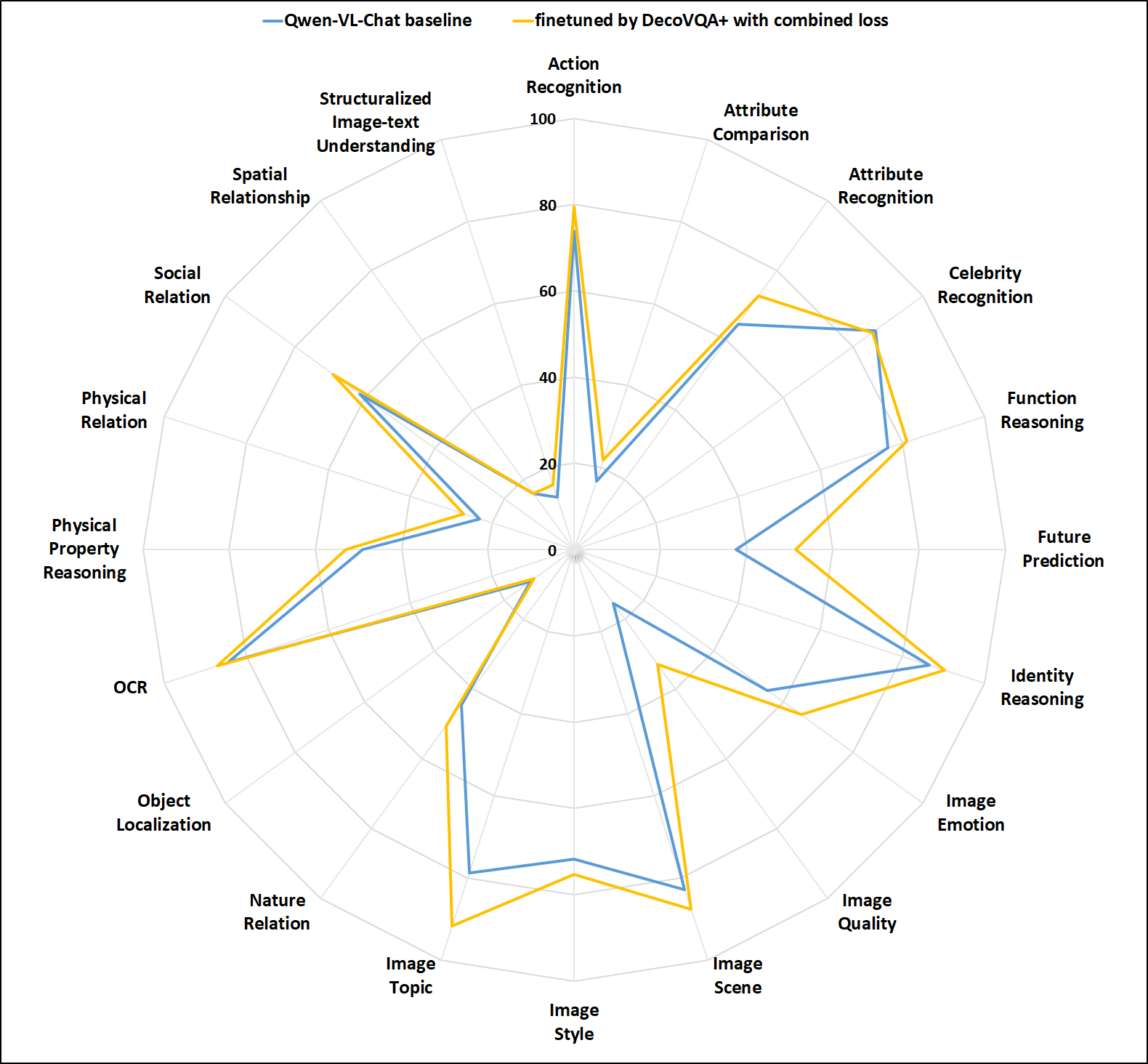}
\caption{Results of different checkpoints of Qwen-VL-Chat across the 20 L-3 ability dimensions defined in MMBench.}
\label{fig:qwen_mmbench}
\end{figure*}

\begin{figure*}[htbp]
\centering
    \includegraphics[width=\textwidth]{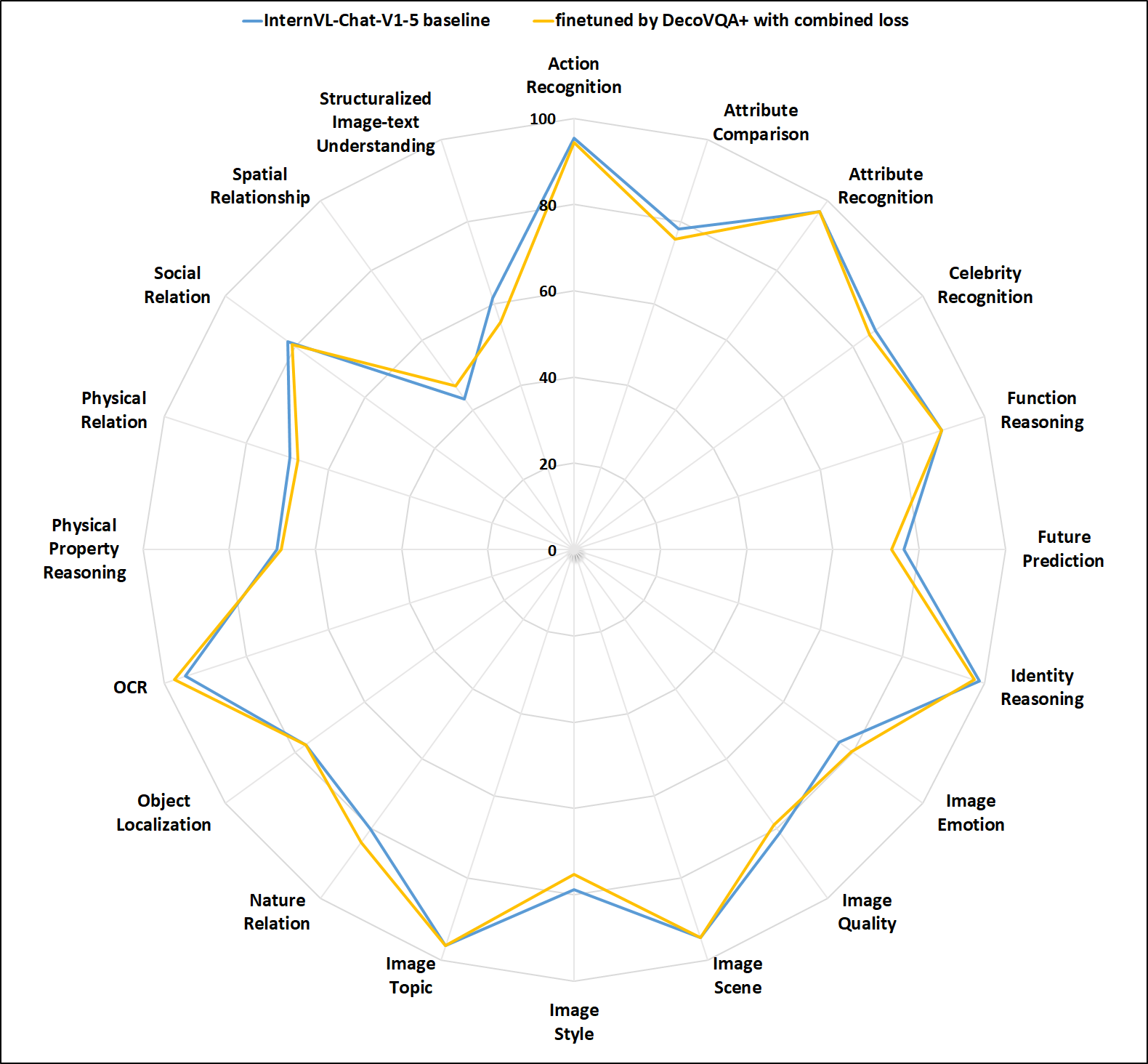}
\caption{Results of different checkpoints of InternVL-Chat-V1-5 across the 20 L-3 ability dimensions defined in MMBench.}
\label{fig:internvl_mmbench}
\end{figure*}

Finetuning may lead to catastrophic forgetting, which hurts the essential all-around performance of MLLMs. MMBench~\citep{liu2023mmbench} is a systematic pipeline that evaluates the comprehensive abilities of MLLMs. Figure~\ref{fig:minigptv2_mmbench}, \ref{fig:llava_mmbench}, \ref{fig:qwen_mmbench} and \ref{fig:internvl_mmbench} demonstrate the evaluation results of different checkpoints on MMBench. It shows that our finetuning does not do harm to most of the abilities, while some of them are even improved after finetuning.

\section{Finetuning with DecoVQA+ vs. with VQA-Introspect}

\label{appendix:dataset_comparison}
The existing public dataset VQA-Introspect has already provided us with complex visual reasoning questions with sub-questions. However, not all questions are complex enough to require decomposition, and a large number of provided sub-questions are repetitive and superficial. To compare with the quality of our proposed dataset, we also finetune MLLMs with the entire training set of VQA-Introspect (excluding the samples used in the evaluation experiments). As shown in Table~\ref{tab:vqd_ability_ft_by_vqaintrospect} and~\ref{tab:ft_by_vqaintrospect}, the performance of the MLLMs finetuned with DecoVQA+ is much better than the ones finetuned with VQA-Introspect. The results demonstrate that the quality of our proposed dataset outperforms the existing public dataset with sub-questions.
\input{figures/vqaft_performance.tex}
\input{tables/ft_by_vqaintrospect}

\section{Comparison with the Unimodal QD Method}
Existing researches~\citep{you-etal-2023-idealgpt, qi-etal-2023-art} tend to use a convincing captioning model to convert images to the language descriptions, and then perform the unimodal question decomposition with LLMs. Table~\ref{tab:compare_language} shows the accuracy gap under the selective VQD inference setting between MLLMs and their corresponding LLMs with GPT-4V as the captioning model. Since critical information in images is often lost during the captioning process, it is very possible for the subsequent inference with QD to fail to answer questions correctly. To sum up, VQD is better than the method "caption + QD".
\input{tables/compare_language_model}

\section{Comparison with In-context Learning Method}
\label{compare_ICL}
Besides finetuning, In-context Learning (ICL) is also a potential approach for VQD. The previous work~\citep{NEURIPS2023_b14cf0a0} has explored VQD based on ICL methods. Therefore, we add an experiment to compare the performance with our finetuning pipeline and with the ICL method.

To fairly compare with the ICL method used in~\citep{NEURIPS2023_b14cf0a0}, we apply the same 2-shot demonstration as the one applied in that paper to decompose questions. The prompt template is shown in~\autoref{fig:prompt_ICL}.

\input{figures/prompt_ICL}

The performance comparison in~\autoref{tab:vqd_compare_ICL} and~\autoref{tab:acc_compare_ICL} shows that the models achieve significantly better performance in VQD ability, VQA accuracy and Whether-to-decompose accuracy through our finetuning pipeline, compared to the ICL method proposed in~\citep{NEURIPS2023_b14cf0a0}.
\input{tables/vqd_compare_ICL}
\input{tables/acc_compare_ICL}

\section{More case studies}
\begin{figure*}[htbp]
    \centering
    \begin{subfigure}[c]{\textwidth}
        \includegraphics[width=\linewidth]{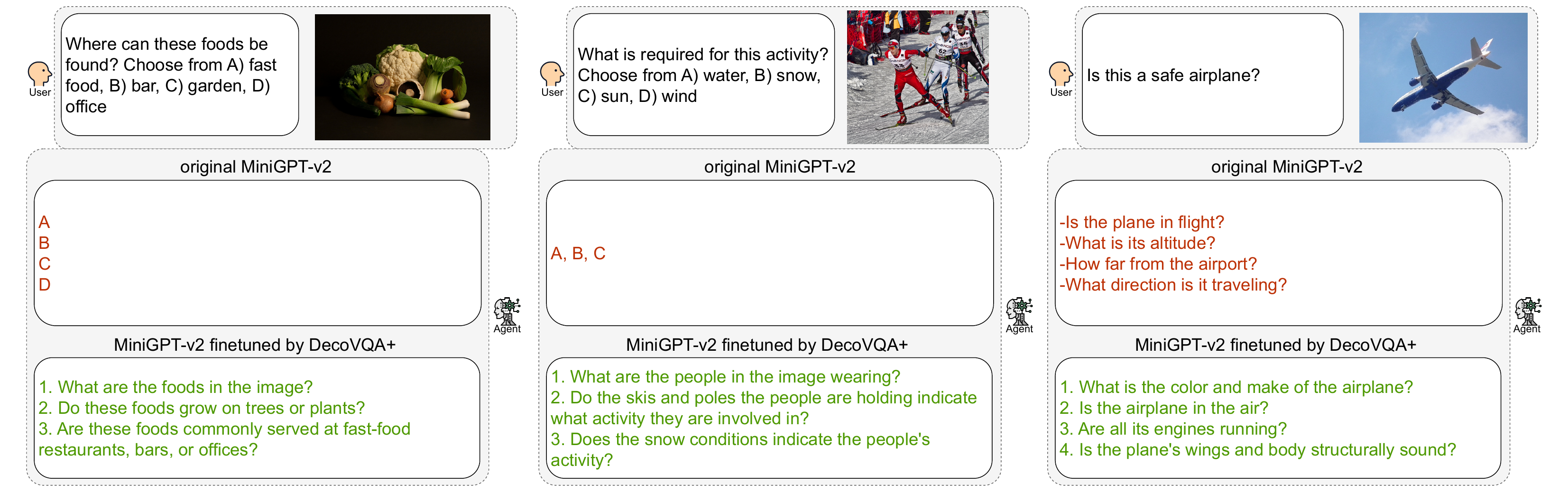}
        \caption{Cases with MiniGPT-v2 before and after being finetuned by DecoVQA+.}
    \end{subfigure}
    \hfill
    \begin{subfigure}[c]{\textwidth}
        \includegraphics[width=\linewidth]{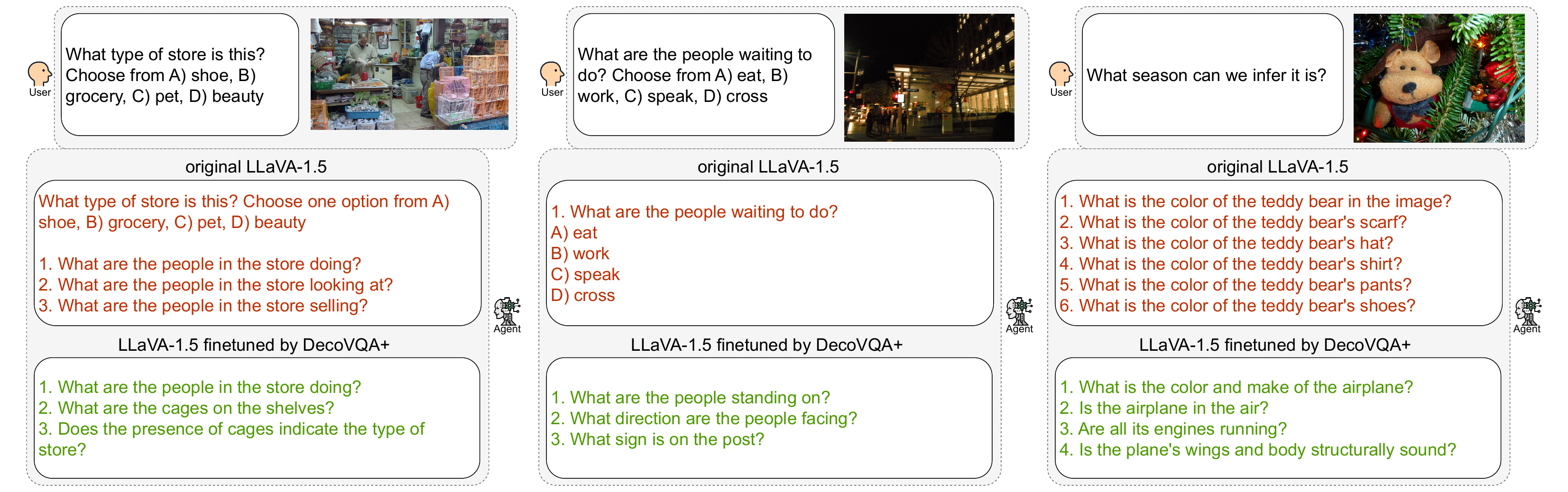}
        \caption{Cases with LLaVA-1.5 before and after being finetuned by DecoVQA+.}
    \end{subfigure}
    \hfill
    \begin{subfigure}[c]{\textwidth}
        \includegraphics[width=\linewidth]{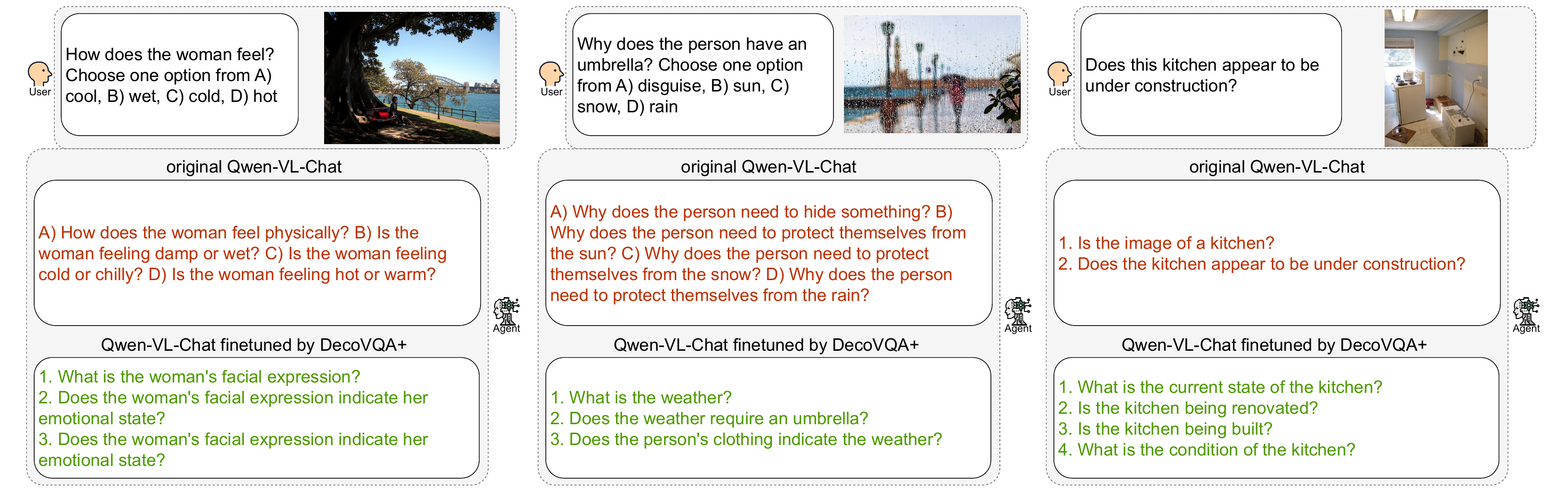}
        \caption{Cases with Qwen-VL-Chat before and after being finetuned by DecoVQA+.}
    \end{subfigure}
    \hfill
    \begin{subfigure}[c]{\textwidth}
        \includegraphics[width=\linewidth]{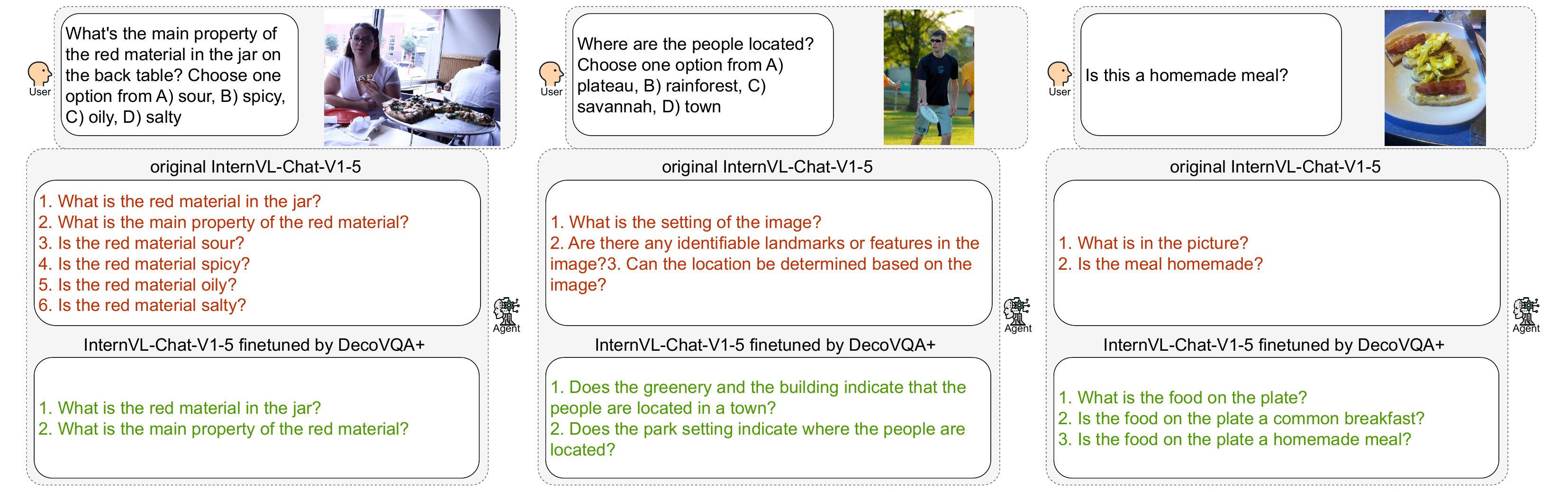}
        \caption{Cases with InternVL-Chat-V1-5 before and after being finetuned by DecoVQA+.}
    \end{subfigure}
    \caption{Case studies showing the comparison of VQD performance by MLLMs before and after finetuning by DecoVQA+.}
    \label{fig:more_case_studies}
\end{figure*}

More case studies in addition to Figure~\ref{eg} are shown in Figure~\ref{fig:more_case_studies}. 

\section{Licensing}
Our proposed datasets SubQuestRater Dataset, DecoVQA, DecoVQA+, and Whether2Deco are built upon the public datasets A-OKVQA and VQA-Introspect. A-OKVQA has the Apache-2.0 License. The licenses of the code for the mentioned MLLMs are listed as follows: MiniGPT-v2 has the BSD-3-Cluase License, LLaVA-1.5 has the Apache-2.0 License, Qwen-VL-Chat has the Tongyi Qianwen License and InternVL-Chat-V1-5 has the MIT License. We publicize all of our proposed datasets and our code under the MIT License.

%% file: tables/alignment_average_score.tex
\begin{table}[htbp]
    \centering
    \resizebox{\linewidth}{!}{
    \begin{tabular}{llccc}
        \toprule
        \textbf{Model} & \textbf{Criteria} & \textbf{GPT-4V} & \textbf{Human} &\textbf{Error Rate} \\
        \midrule
         & \textbf{Non-Repetition} & 51.92 & 52.37 & -0.86\%\\
         
        \makecell[l]{\textbf{Original}\\\textbf{MiniGPT-v2}} & \textbf{Relevance} & 40.86 & 39.34 & 3.86\%\\
         & \textbf{Groundedness} & 47.65 & 47.84 & -0.40\%\\
         
        \noalign{\vskip 1ex}\cdashline{1-5}\noalign{\vskip 1ex}
        
         & \textbf{Non-Repetition} & 94.22 & 93.79 & 0.46\%\\
        \makecell[l]{\textbf{MiniGPT-v2}\\\textbf{finetuned by DecoVQA}}& \textbf{Relevance} & 74.54 & 75.79 & -1.65\%\\ 
         & \textbf{Groundedness} & 86.36 & 87.47 & -1.27\%\\
        
        \bottomrule
    \end{tabular}
    }

    \caption{Comparison of average scores of the judgements on SubQuestRater dataset from GPT-4V and human reviewers on three criteria. We regard judgements from human reviewers as the ground truth when computing the error rate.
    }
    \label{tab:alignment_average_score}
\end{table}

%% file: tables/alignment_pearson_spearman.tex
\begin{table}[htbp]
    \centering
    \resizebox{\linewidth}{!}{
    \begin{tabular}{llcc}
        \toprule
        \textbf{Model} & \textbf{Criteria} & \textbf{Pearson} & \textbf{Spearman}\\
        \midrule

         & \textbf{Non-Repetition} & 0.828 & 0.820\\
        \makecell[l]{\textbf{Original}\\\textbf{MiniGPT-v2}}& \textbf{Relevance} & 0.813 & 0.813\\ 
         & \textbf{Groundedness} & 0.801 & 0.795\\
         
        \noalign{\vskip 1ex}\cdashline{1-4}\noalign{\vskip 1ex}
         
         & \textbf{Non-Repetition} & 0.867 & 0.864\\
        \makecell[l]{\textbf{MiniGPT-v2}\\\textbf{finetuned by DecoVQA}}& \textbf{Relevance} & 0.804 & 0.784\\ 
         & \textbf{Groundedness} & 0.832 & 0.886\\
        
        \bottomrule
    \end{tabular}
    }

    \caption{Pearson and Spearman correlation coefficients of the judgements on SubQuestRater dataset from GPT-4V and human reviewers on three criteria. All results are statistically highly significant (p-value < 0.001).
    }
    \label{tab:alignment_pearson_spearman}
\end{table}

%% file: figures/4modelsperformance.tex
\begin{table}[htbp]
    \centering
    \resizebox{\linewidth}{!}{
    \begin{tabular}{lcccc}
        \toprule
        \textbf{MiniGPT-v2} & \textbf{DecoVQA+100} & \textbf{DecoVQA+200} & \textbf{DecoVQA+400}\\
        
        \noalign{\vskip 1ex}\cdashline{1-5}\noalign{\vskip 1ex}
        
        \textbf{Non-Repetition} &\textbf{91.56} 
 &91.48 & 88.35\\
        \textbf{Relevance} &70.19 
 &\textbf{72.59} &71.64 \\
        \textbf{Groundedness} &\textbf{87.35} &85.49 & 83.15\\

        \midrule

        \textbf{LLaVA-1.5} & \textbf{DecoVQA+100} & \textbf{DecoVQA+200} & \textbf{DecoVQA+400}\\
        
        \noalign{\vskip 1ex}\cdashline{1-5}\noalign{\vskip 1ex}
        
        \textbf{Non-Repetition} & 71.92 & 88.97 & \textbf{94.18} \\
        \textbf{Relevance} & 67.92 & \textbf{79.41} & 78.67 \\
        \textbf{Groundedness} & 80.94 & \textbf{89.28} & 85.63 \\
        
        \bottomrule
    \end{tabular}
    }
    \caption{Ablation study about VQD ability on finetuning models with DecoVQA+ with a varying sample number. DecoVQA+400 is the version of DecoVQA+ with which we finetune MLLMs in other experiments.
    }
    \label{4models}
\end{table}

%% file: tables/ablation_50_100_200.tex
\begin{table}[htbp]
    \centering
    \resizebox{\linewidth}{!}{
    \begin{tabular}{lllll}
        \toprule
        \textbf{Models} & \textbf{A-OKVQA} & \textbf{GQA} & \textbf{VQA-Introspect} & \textbf{Whether2Deco}\\
        \midrule
        
        \textbf{MiniGPT-v2} & 41.2 & 44.2 & 62.1 & 46.8\\
        \quad finetuned by DecoVQA+100 & 59.3 \up{$\uparrow$ (+18.1)} & \textbf{51.4} \up{$\uparrow$ (+7.2)} & 70.1 \up{$\uparrow$ (+8.0)} & 54.3 \up{$\uparrow$ (+7.5)}\\
        \quad finetuned by DecoVQA+200 & \textbf{61.3} \up{$\uparrow$ (+20.1)} & 50.9 \up{$\uparrow$ (+6.7)} & \textbf{73.4} \up{$\uparrow$ (+11.3)} & \textbf{63.8} \up{$\uparrow$ (+17.0)}\\
        \quad finetuned by DecoVQA+400 & 60.7 \up{$\uparrow$ (+19.5)} & 50.7 \up{$\uparrow$ (+6.5)} & 72.1 \up{$\uparrow$ (+10.0)} & 61.0 \up{$\uparrow$ (+14.2)}\\

        \noalign{\vskip 1ex}\cdashline{1-5}\noalign{\vskip 1ex}
        
        \textbf{LLaVA-1.5} & 67.7 & 52.1 & 67.2 & 49.3\\
        \quad finetuned by DecoVQA+100 & 73.4 \up{$\uparrow$ (+5.7)} & 53.6 \up{$\uparrow$ (+1.5)} & 72.5 \up{$\uparrow$ (+5.3)} & 56.0 \up{$\uparrow$ (+6.7)}\\
        \quad finetuned by DecoVQA+200 & \textbf{74.4} \up{$\uparrow$ (+6.7)} & \textbf{57.9} \up{$\uparrow$ (+5.8)} & \textbf{78.7} \up{$\uparrow$ (+11.5)} & \textbf{69.0} \up{$\uparrow$ (+19.7)}\\
        \quad finetuned by DecoVQA+400 & 72.7 \up{$\uparrow$ (+5.0)} & 57.2 \up{$\uparrow$ (+5.1)} & 75.4 \up{$\uparrow$ (+8.2)} & 68.8 \up{$\uparrow$ (+19.5)}\\
        \bottomrule
    \end{tabular}
    }
    \caption{Ablation study about VQA accuracy and Whether2Deco accuracy on finetuning models with DecoVQA+ with a varying sample number. DecoVQA+400 is the version of DecoVQA+ with which we finetune MLLMs in other experiments.}
    \label{tab:ablation_ar_loss}
\end{table}

%% file: tables/dataset_overview.tex
\begin{table*}[htbp]
    \centering
    \begin{tabular}{lllrr}
        \toprule
        \textbf{Dataset} & \textbf{Dataset Type} & \textbf{Question Type} & \textbf{\# Images} & \textbf{\# Questions}\\
        \midrule 
        A-OKVQA & external knowledge & multiple choice & 6,030 & 6,702\\
        GQA & visual reasoning & open-ended questions & 398 & 12,578\\
        VQA-Introspect & visual reasoning & open-ended questions & 17,495& 22,793\\
        \bottomrule
    \end{tabular}
    \caption{Experimental statistics for pubilc datasets used in the paper.}
    \label{tab:dataset}
\end{table*}

%% file: tables/proposed_dataset_overview.tex
\begin{table*}[htbp]
    \centering
    \begin{tabular}{lllrr}
        \toprule
        \textbf{Dataset} & \textbf{Usage} & \textbf{Motivation} & \textbf{\# Images} & \textbf{\# Questions}\\
        \midrule
        SubQuestRater & evaluation & measuring the quality of sub-questions & 200 & 200\\
        DecoVQA & finetuning & \makecell[l]{improving the VQD ability} & 397* & 400\\
        DecoVQA+ & finetuning & \makecell[l]{improving the selective VQD ability} & 397* & 400\\
        Whether2Deco & evaluation & \makecell[l]{testing the models' ability to identify \\ whether a question requires decomposition} & 395*& 400\\
        \bottomrule
    \end{tabular}
    \caption{Experimental statistics for proposed datasets in the paper. *Several images correspond to more than one question.}
    \label{tab:proposed_dataset}
\end{table*}

%% file: figures/criteria.tex
\begin{algorithm}[htbp]
    \DontPrintSemicolon
    \SetKwInOut{Sub}{$\boldsymbol{q}$}
    \SetKwInOut{Subs}{$\boldsymbol{\mathbb{Q}}$}
    \SetKwInOut{BScores}{$\boldsymbol{b_1}, \boldsymbol{b_2}, \boldsymbol{b_3}$}
    \SetKwInOut{BScoreSets}{$\boldsymbol{B_1}, \boldsymbol{B_2}, \boldsymbol{B_3}$}
    \SetKwInOut{FinalScores}{$\boldsymbol{s_1}, \boldsymbol{s_2}, \boldsymbol{s_3}$}
    
    \Sub{Sub-question}
    \Subs{Set of sub-questions from one sample}
    \BScores{Binary score for 3 criteria of a sub-question}
    \BScoreSets{Lists of binary scores for 3 criteria of sub-questions}
    \FinalScores{Score for 3 criteria for a sample}
    
    \emph{Check if there are effective sub-questions}\;

    \If{$\mathbb{Q}==\emptyset$}{
        $s_1$ = 0\\
        $s_2$ = 0\\
        $s_3$ = 0\\
    }
    \Else{
    \For{q in $\mathbb{Q}$}{
    $b_1, b_2, b_3=\{ScoreModel(q) \mid q\subset \boldsymbol{\mathbb{Q}}$\} \\
    \For{$i$ $\in$ [1,3]}{
    $AppendtoList(b_i,\boldsymbol{B_i})$
    }
    }
    \For{$j$ $\in$ [1,3]}{
    $s_i=CalculateAverage(\boldsymbol{B_j})$
    }
    }
    \Return{$s_1, s_2, s_3$}
    \caption{Evaluation algorithm for the quality of sub-questions}
    \label{algorithm_eva}
\end{algorithm}

%% file: figures/ablation.tex
\begin{table*}[htbp]
    \centering
    \resizebox{\linewidth}{!}{
    \begin{tabular}{lcccc}
        \toprule
        \textbf{MiniGPT-v2}& \textbf{original Model} & \textbf{finetuned by DecoVQA} & \textbf{finetuned by DecoVQA+} & \textbf{finetuned by DecoVQA+ with SelectiveVQD Loss}\\
        \noalign{\vskip 1ex}\cdashline{1-5}\noalign{\vskip 1ex}
        \textbf{Non-Repetition} & 47.52 & \textbf{93.72} & 88.35 & 90.58 \\
        \textbf{Relevance} & 36.65 & \textbf{74.17} & 71.64 & 73.73 \\
        \textbf{Groundedness} & 43.30 & \textbf{85.98} & 83.15 & 84.53 \\
        \midrule
        \textbf{LLaVA-1.5}& \textbf{original Model} & \textbf{finetuned by DecoVQA} & \textbf{finetuned by DecoVQA+} & \textbf{finetuned by DecoVQA+ with SelectiveVQD Loss}\\
        \noalign{\vskip 1ex}\cdashline{1-5}\noalign{\vskip 1ex}
        \textbf{Non-Repetition} & 42.19 & 92.04 & \textbf{94.18} & 92.68 \\
        \textbf{Relevance} & 37.33 & \textbf{81.62} & 78.67 & 78.48 \\
        \textbf{Groundedness} & 44.17 & \textbf{86.19} & 85.63 & 84.39 \\
        \midrule
        \textbf{Qwen-VL-Chat}& \textbf{original Model} & \textbf{finetuned by DecoVQA} & \textbf{finetuned by DecoVQA+} & \textbf{finetuned by DecoVQA+ with SelectiveVQD Loss}\\
        \noalign{\vskip 1ex}\cdashline{1-5}\noalign{\vskip 1ex}
        \textbf{Non-Repetition} & 32.10 & 80.66 & 89.03 & \textbf{89.15} \\
        \textbf{Relevance} & 27.15 & \textbf{69.52} & 68.73 & 67.15 \\
        \textbf{Groundedness} & 26.49 & 77.34 & \textbf{78.92} & 77.51 \\
        \midrule
        \textbf{InternVL-Chat-V1-5}& \textbf{original Model} & \textbf{finetuned by DecoVQA} & \textbf{finetuned by DecoVQA+} & \textbf{finetuned by DecoVQA+ with SelectiveVQD Loss}\\
        \noalign{\vskip 1ex}\cdashline{1-5}\noalign{\vskip 1ex}
        \textbf{Non-Repetition} & 82.41 & 87.40 & 92.76 & \textbf{94.11} \\
        \textbf{Relevance} & 73.42 & 81.11 & \textbf{83.38} & 83.30 \\
        \textbf{Groundedness} & 78.01 & 87.62 & \textbf{90.15} & 89.47 \\
        \bottomrule
    \end{tabular}
    }

    \caption{Comparison of VQD abilities of all the original models and their corresponding finetuned versions.
    }
    \label{ablation_100200}
\end{table*}

%% file: figures/eval_mc.tex
\begin{algorithm}[htbp]
    \DontPrintSemicolon
    \SetKwInOut{answer}{$\boldsymbol{a}$}
    \SetKwInOut{mappedAnswer}{$\boldsymbol{\hat{a}}$}
    \SetKwInOut{n}{$\boldsymbol{n}$}
    \SetKwInOut{threshold}{$\boldsymbol{\tau}$}
    
    \answer{Model answer}
    \mappedAnswer{Mapped option}
    \n{Number of exact match options}
    \threshold{Threshold of sentence length}
    
    \emph{Attempt exact match and get n mentioned options from model answer}\;
    \If{$n==1$}{
        $\hat{a}=$ ExactMatch$(a)$\;
    }
    \Else{
        \If{$len(tokenize(a))\leq\tau$}{
            $\hat{a} =$ SentenceTransformer$(a)$\;
        }
        \Else{
            $\hat{a} =$ GPT-4$(a)$\
        }
    } 
    \Return{$\hat{a}$}    
    \caption{Robust algorithm for measuring accuracy on MC datasets}
    \label{algorithm_mc}
\end{algorithm}

%% file: figures/vqaft_performance.tex
\begin{table}[!htb]
    \centering
    \resizebox{\linewidth}{!}{
    \begin{tabular}{lccc}
        \toprule
        \textbf{MiniGPT-v2} & \textbf{Finetuned by VQAintrospect} & \textbf{Finetuned by DecoVQA+} \\
        \noalign{\vskip 1ex}\cdashline{1-4}\noalign{\vskip 1ex}

        \textbf{Non-Repetition} & 17.20 & \textbf{88.35} \\
        \textbf{Relevance} & 13.08 & \textbf{71.64} \\
        \textbf{Groundedness} & 14.87 & \textbf{83.15} \\
        
        \midrule

        \textbf{LLaVA-1.5} & \textbf{Finetuned by VQAintrospect} & \textbf{Finetuned by DecoVQA+} \\
        \noalign{\vskip 1ex}\cdashline{1-4}\noalign{\vskip 1ex}        
        \textbf{Non-Repetition} & 21.52 & \textbf{94.18} \\
        \textbf{Relevance} & 76.90* & \textbf{78.67} \\
        \textbf{Groundedness} & \textbf{93.50*} & 85.63 \\
        
        \bottomrule
    \end{tabular}
    }
    \caption{Comparison of VQD abilities on MLLMs before and after finetuning with VQA-Introspect and with DecoVQA+. *Here for most of the original questions, LLaVA-1.5 produces one high quality sub-question, then repeats it for 2-3 times, causing relatively high score on Relevance and Groundedness, yet very low in Non-Repetition score. 
    }
    \label{tab:vqd_ability_ft_by_vqaintrospect}
\end{table}

%% file: tables/ft_by_vqaintrospect.tex
\begin{table}[!htb]
    \centering
    \resizebox{\linewidth}{!}{
    \begin{tabular}{lllll}
        \toprule
        \textbf{Models} & \textbf{A-OKVQA} & \textbf{GQA} & \textbf{VQA-Introspect} & \textbf{Whether2Deco}\\
        \midrule
         
        \textbf{MiniGPT-v2} & 41.2 & 44.2 & 62.1 & 46.8\\
        \quad finetuned by VQAIntrospect & 48.8 \up{$\uparrow$ (+7.6)} & 39.6 \down{$\downarrow$ (-4.6)} & 63.7 \up{$\uparrow$ (+1.6)} & 37.3 \down{$\downarrow$ (-9.5)}\\
        \quad finetuned by DecoVQA+ & \textbf{60.7} \up{$\uparrow$ (+19.5)} & \textbf{50.7} \up{$\uparrow$ (+6.5)} & \textbf{72.1} \up{$\uparrow$ (+10.0)} & \textbf{61.0} \up{$\uparrow$ (+14.2)}\\
        \noalign{\vskip 1ex}\cdashline{1-5}\noalign{\vskip 1ex}
        
        \textbf{LLaVA-1.5} & 67.7 & 52.1 & 67.2 & 49.3\\
        \quad finetuned by VQAIntrospect & 68.4 \up{$\uparrow$ (+0.7)} & 51.8 \down{$\downarrow$ (-0.3)} & \textbf{81.1} \up{$\uparrow$ (+13.9)} & 4.8* \down{$\downarrow$ (-44.5)}\\
        \quad finetuned by DecoVQA+ & \textbf{72.7} \up{$\uparrow$ (+5.0)} & \textbf{57.2} \up{$\uparrow$ (+5.1)} & 75.4 \up{$\uparrow$ (+8.2)} & \textbf{68.8} \up{$\uparrow$ (+19.5)}\\
        \bottomrule
    \end{tabular}
    }
    \caption{Comparison of VQA accuracy (\%) on external knowledge (A-OKVQA) and visual reasoning (GQA and VQA-Introspect) datasets and Whether2Deco accuracy (\%) before and after fine-tuning MLLMs with VQA-Introspect and with DecoVQA+. *Here LLaVA-1.5 fails to follow the pre-defined answering template, but to perform pure question decomposition instead of selective decomposition.
    }
    \label{tab:ft_by_vqaintrospect}
\end{table}

%% file: tables/compare_language_model.tex
\begin{table}[htbp]
    \centering
    \begin{tabular}{ll}
        \toprule
        \textbf{Models} & \textbf{VQA-Introspect}\\
        \midrule
        
        \textbf{MiniGPT-v2} & 62.1\\
        Llama2-Chat-7B-HF & 46.2\\
       
        \noalign{\vskip 1ex}\cdashline{1-2}\noalign{\vskip 1ex}
        
        \textbf{LLaVA-1.5} & 67.2\\
        Vicuna-13B-v1.5 & 62.3\\
        \bottomrule
    \end{tabular}
    
    \caption{Comparison of VQA accuracy (\%) between MLLMs and their corresponding language models on VQA-Introspect.}
    \label{tab:compare_language}
\end{table}

%% file: figures/prompt_ICL.tex
\begin{figure*}[htbp]

\begin{tcolorbox}[title=Prompt under ICL setting (two-shot)]
Please firstly decompose the given question into several image-relevant sub-questions to help you answer the given question. Please avoid giving repeated sub-questions or generating an excessive number. Feel free to suggest an appropriate quantity based on your judgment. Here are two examples you can follow to decompose the question:\\
\\
Example 1\\
Question: Is the banana ripe enough to eat?\\
Sub-questions: 1. Is the banana yellow?\\
\\
Example 2\\
Question: Is it cold outside?\\
Sub-questions: 1. Are any people wearing jackets?\\
\\
Input\\
Question: \{question\}\\
Sub-questions:
\end{tcolorbox}
\caption{Prompt under ICL setting (two-shot)}
\label{fig:prompt_ICL}
\end{figure*}

%% file: tables/vqd_compare_ICL.tex
\begin{table}[htbp]
    \centering
    \resizebox{\linewidth}{!}{
    \begin{tabular}{lccc}
        \toprule
        \textbf{Model} & \textbf{Non-Repetition} & \textbf{Relevance} & \textbf{Groundedness}\\
        \midrule
        \textbf{MiniGPT-v2 (zero-shot)} & 47.52 & 36.65 & 43.30\\
        \textbf{MiniGPT-v2 (ICL)} & 54.65 & 49.64 & 49.97\\
        \textbf{MiniGPT-v2 (finetuned by DecoVQA+)} & \textbf{88.35} & \textbf{71.64} & \textbf{83.15}\\
        \noalign{\vskip 1ex}\cdashline{1-4}\noalign{\vskip 1ex}
        \textbf{LLaVA-1.5 (zero-shot)} & 42.19 & 37.33 & 44.17\\
        \textbf{LLaVA-1.5 (ICL)} & 69.45 & 65.32 & 62.58\\
        \textbf{LLaVA-1.5 (finetuned by DecoVQA+)} & \textbf{94.18} & \textbf{78.67} & \textbf{85.63}\\
        
        \bottomrule
    \end{tabular}
    }
    \caption{Comparison of VQD ability between MLLMs finetuned by DecoVQA+ and inference with ICL-method across three evaluation criteria.}
    \label{tab:vqd_compare_ICL}
\end{table}

%% file: tables/acc_compare_ICL.tex
\begin{table}[htbp]
    \centering
    \resizebox{\linewidth}{!}{
    \begin{tabular}{lcccc}
        \toprule
        \textbf{Model} & \textbf{A-OKVQA} & \textbf{GQA} & \textbf{VQA-Introspect} & \textbf{Whether2Deco}\\
        \midrule
        \textbf{MiniGPT-v2 (zero-shot)} & 41.2 & 44.2 & 62.1 & 46.8\\
        \textbf{MiniGPT-v2 (ICL)} & 40.1 & 43.6 & 60.5 & 46.8\\
        \textbf{MiniGPT-v2 (finetuned by DecoVQA+)} & \textbf{64.0} & \textbf{51.7} & \textbf{72.5} & \textbf{71.5}\\
        \noalign{\vskip 1ex}\cdashline{1-5}\noalign{\vskip 1ex}
        \textbf{LLaVA-1.5 (zero-shot)} & 67.7 & 52.1 & 67.2 & 49.3\\
        \textbf{LLaVA-1.5 (ICL)} & 65.1 & 51.3 & 67.4 & 49.3\\
        \textbf{LLaVA-1.5 (finetuned by DecoVQA+)} & \textbf{73.9} & \textbf{56.7} & \textbf{75.8} & \textbf{75.0}\\
        
        \bottomrule
    \end{tabular}
    }
    \caption{Comparison of Accuracy (\%) between MLLMs finetuned by DecoVQA+ and inference with ICL-method.}
    \label{tab:acc_compare_ICL}
\end{table}